\documentclass{article}

\usepackage{amssymb}
\usepackage{amsmath}
\usepackage{dsfont}
\usepackage{xcolor}
\usepackage{todonotes}
\usepackage{url}
\usepackage{subfig}
\usepackage{authblk}

\title{Generative adversarial networks and adversarial methods in biomedical image analysis}
\date{}
\author[1]{Jelmer M. Wolterink}
\author[2]{Konstantinos Kamnitsas}
\author[3]{Christian Ledig}
\author[1]{Ivana~I\v{s}gum}
\affil[1]{Image Sciences Institute, University Medical Center Utrecht, Utrecht, The Netherlands}
\affil[2]{Imperial College London, London, United Kingdom}
\affil[3]{Imagen Technologies, New York, New York, United States of America}
\begin{document}

\maketitle

\begin{abstract}
Generative adversarial networks (GANs) and other adversarial methods are based on a game-theoretical perspective on joint optimization of two neural networks as players in a game. Adversarial techniques have been extensively used to synthesize and analyze biomedical images. We provide an introduction to GANs and adversarial methods, with an overview of biomedical image analysis tasks that have benefited from such methods. We conclude with a discussion of strengths and limitations of adversarial methods in biomedical image analysis, and propose potential future research directions.
\end{abstract}

\section{Introduction}
Recent years have seen a strong increase in deep learning applications to medical image analysis \cite{Litj17}. Many of these applications are supervised, consisting of a convolutional neural network (CNN) that is optimized to provide a desired prediction given an input image. For example, given a medical image, we may require the CNN to obtain a segmentation of a number of anatomical structures. 
These are \textit{discriminative} models, in which the CNN tries to discriminate between images or image voxels that correspond to different classes. 
Optimization of this CNN is \textit{supervised} by a loss function that quantifies the agreement between model predictions and reference labels. Convolutional layers and downsampling layers are used to discard redundant information from the input, obtain invariance to e.g. translations, and transform a high-dimensional input into a low-dimensional prediction.

In contrast to discriminative models, \textit{generative models} aim to learn the underlying distribution of the data and the generative process that creates them. They can thus be used to obtain an understanding of the structure of the data, or to generate new data by sampling from the model. As an example, a low-dimensional input, such as a noise vector or a vector that encodes the required characteristics of the output, can be transformed into a high-dimensional output, such as an image. Generative adversarial networks (GANs) have recently emerged as a powerful class of generative models \cite{Good14}. Central to the idea of GANs is the joint optimization of two neural networks with opposing goals, i.e. \textit{adversarial} training. The first network is a generator network that maps input from a source domain, which is often low-dimensional, to a target domain such as the high-dimensional space of natural images. This network is jointly optimized with a second, adversarial network, called the discriminator network. The generator tries to generate outputs that the discriminator network cannot distinguish from a dataset of real examples. Both the generator and discriminator are optimized based on the output of the discriminator: if the discriminator can easily distinguish the generator's outputs from samples in the real dataset, the weights of the generator need to be adjusted accordingly.

It has been shown that the training of discriminative or regression models such as segmentation CNNs can also benefit from the signal of an adversarial network. For example, an adversarial network could provide a loss term to a segmentation CNN that quantifies to what extent its segmentation outputs are similar to real image segmentations. 
This is a task for which it is challenging to hand-craft an appropriate loss function. Adversarial networks can enable the prediction of topologically more reasonable segmentations, e.g. segmentation maps without holes or fragments.
This type of adversarial training has found its way to many applications in biomedical image analysis. 

We provide an introduction to GANs and adversarial methods with a focus on applications in biomedical image analysis. In Sec. \ref{sec:GANS} we describe GANs and methods to optimize GANs. In Sec. \ref{sec:advmethods} we describe the use of adversarial networks to map images from one domain to another domain, which could benefit many medical image analysis techniques such as segmentation, modality synthesis and artifact reduction. Sec. \ref{sec:theory_da} provides an introduction to domain adaptation with adversarial methods. Sec. \ref{sec:GANSinMIA} describes a number of applications of GANs and adversarial methods in biomedical image analysis. Sec. \ref{sec:discuss} provides a discussion of some of the current strengths and limitations of GANs and adversarial methods in biomedical image analysis, as well as a discussion of potential future research directions.

\begin{figure}
\centering
\includegraphics[width=0.8\textwidth]{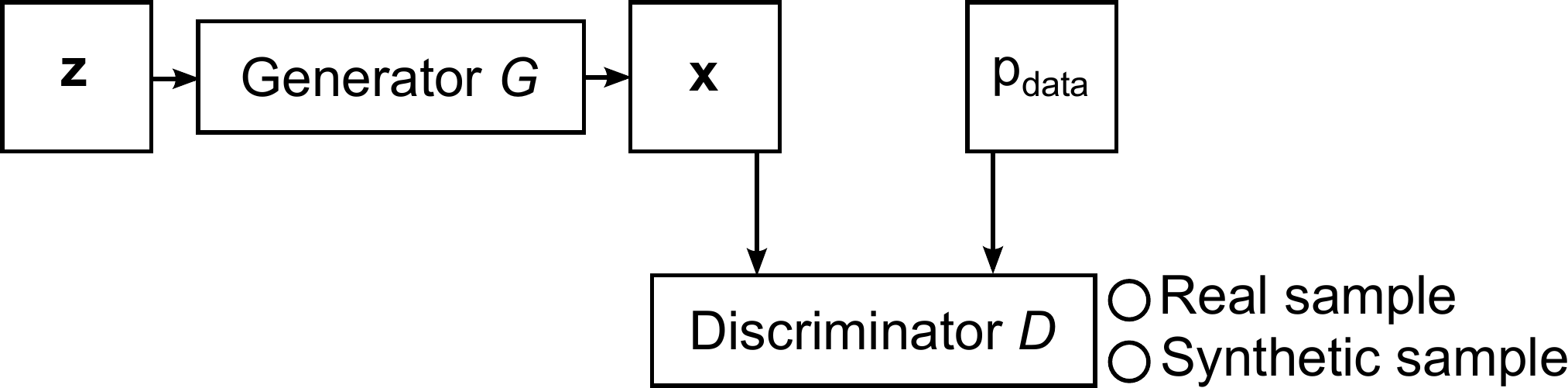}
\caption{Generative adversarial network. The generator $G$ takes a noise vector $\mathbf{z}$ sampled from a distribution $p_z$ as input and uses fully connected or convolutional layers to transform this vector into a sample $\mathbf{x}$. The discriminator $D$ tries to distinguish these samples from samples drawn from the real data distribution $p_{\text{data}}$.}
\label{fig:gan}
\end{figure}

\section{Generative adversarial networks}
\label{sec:GANS}
Generative adversarial networks consist of two neural networks \cite{Good14}. The first network, the \textit{generator}, tries to generate synthetic but perceptually convincing samples $\mathbf{x} \in p_{\text{fake}}$ that appear to have been drawn from a real data distribution $p_{\text{data}}$. It transforms noise vectors $\mathbf{z}$ drawn from a distribution $p_z$ into new samples, i.e. $\mathbf{x}=G(\mathbf{z})$ (Fig. \ref{fig:gan}). The second network, the \textit{discriminator}, has access to real samples from $p_{\text{data}}$ and to the samples generated by $G$, and tries to discriminate between these two. GANs are trained by solving the following optimization problem that the discriminator is trying to maximize and the generator is trying to minimize.

\begin{equation}
\underset{G}{\text{min}}~\underset{D}{\text{max}}~V(D,G)=\underset{\mathbf{x}\sim p_{\text{data}}}{\mathds{E}} [\log{D(\mathbf{x})}]+\underset{\mathbf{z}\sim p_z}{\mathds{E}} [\log{(1-D(G(\mathbf{z})))}],
\label{eq:gan}
\end{equation}

where $G$ is the generator, $D$ is the discriminator, $V(D,G)$ is the objective function, $p_{\text{data}}$ is the distribution of real samples and $p_z$ is a distribution from which noise vectors are drawn, e.g. a uniform distribution or spherical Gaussian. The final layer of the discriminator network contains a sigmoid activation function, so that $D(\mathbf{x}), D(G(\mathbf{z}))\in [0,1]$. By maximizing the value function, the discriminator minimizes the error of its predictions with respect to target values 1 and 0 for real and fake samples, respectively. Conversely, the generator tries to minimize the chance that the discriminator will predict a 0 for fake samples. Hence, the loss of the generator depends directly on the performance of the discriminator. 

\begin{figure}
\centering
\includegraphics[width=0.9\textwidth]{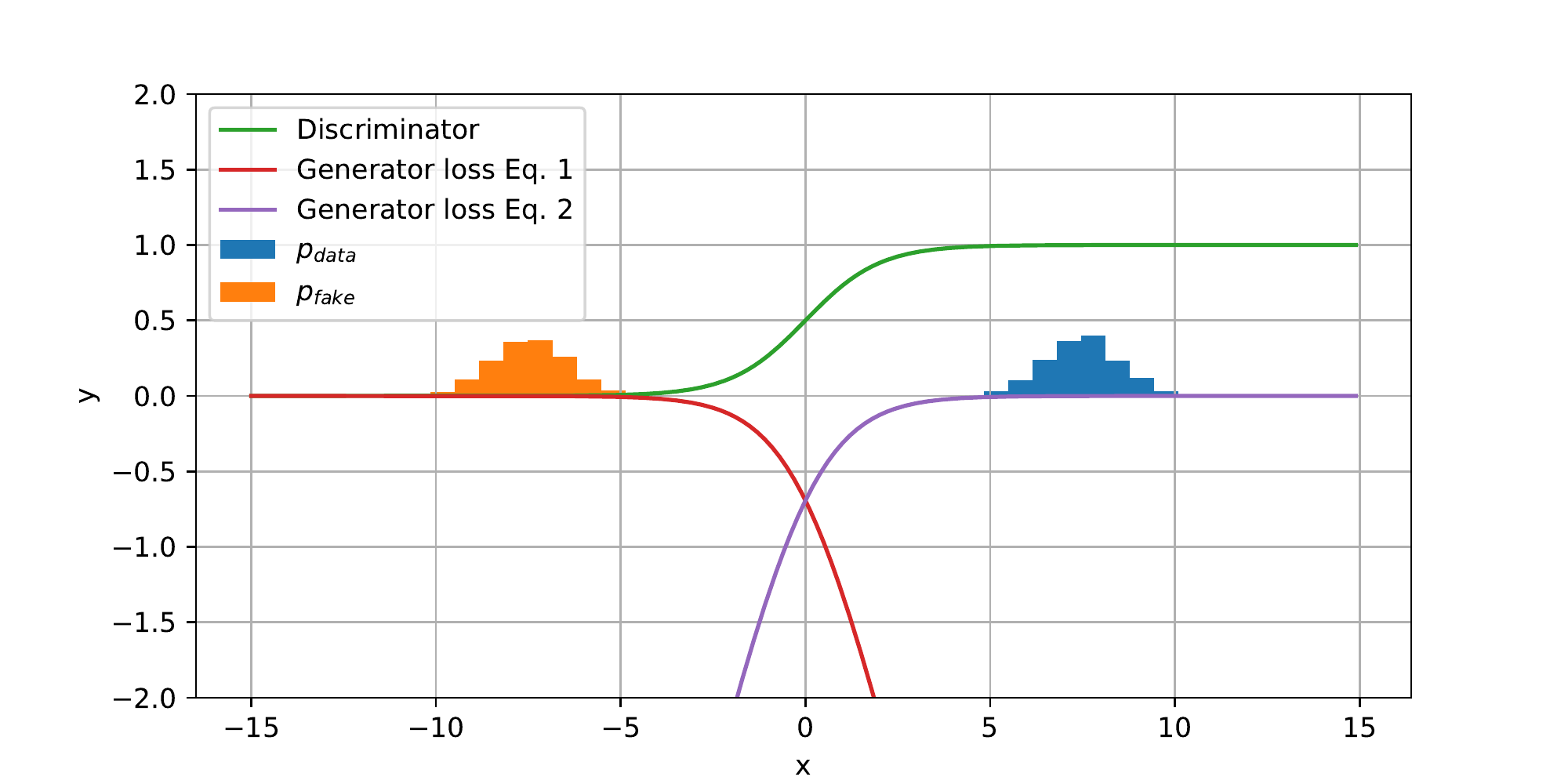}
\caption{Objective function on real and fake data. When using the generator loss in Eq. \protect{\ref{eq:gan}}, gradients for fake samples saturate when real and fake data are easily separated. When using the alternative loss in Eq. \protect{\ref{eq:ganma}}, gradients for fake samples do not saturate.}
\label{fig:objectives}
\end{figure}

\subsection{Objective functions}
\label{sec:objectives}
The iterative approach when training GANs, i.e. finding a saddle point in Eq. \ref{eq:gan}, tends to be unstable. This makes the optimization process challenging in practice. 
Optimization of the generator depends directly on gradients provided by the discriminator for synthetic samples. A problem arises when the discriminator can easily distinguish samples in $p_{\text{fake}}$ from those in $p_{\text{data}}$, as is common at the beginning of GAN training. In that case, the gradient for $\log{(1-D(G(\mathbf{z})))}$ is close to zero (Fig. \ref{fig:objectives}). Consequently, the generator will fail to update its parameters and minimize its loss. Therefore, an alternative loss function for the generator is used in practice \cite{Good14}:

\begin{equation}
\underset{G}{\text{min}}~V(D,G)=-\underset{\mathbf{z}\sim p_z}{\mathds{E}} [\log{D(G(\mathbf{z}))}].
\label{eq:ganma}
\end{equation}

While in Eq. \ref{eq:gan} the objective for the generator is to minimize the probability that the discriminator identifies generated samples as fake, in Eq. \ref{eq:ganma} the objective is to maximize the probability that the discriminator identifies generated samples as real. Fig. \ref{fig:objectives} shows how this objective has a strong gradient for fake samples that are far from the discriminator's decision boundary, and how this is not the case when using the objective in Eq. \ref{eq:gan}.

\begin{table}[tp]
\caption{Discriminator and generator objectives in different GAN variants: original GAN (GAN, Eq. \ref{eq:gan}), an alternative in which the generator maximizes the probability that its samples are identified as real (GAN-MA, Eq. \ref{eq:ganma}), least-squares GAN (LSGAN), Wasserstein GAN (WGAN, Eq. \ref{eq:earthmover}), and  Wasserstein GAN with gradient penalty (WGAN-GP, Eq. \ref{eq:gradientpenalty}). In all cases, the discriminator or generator is trying to minimize the listed loss.}
\resizebox{\textwidth}{!}{
\begin{tabular}{lll}
& Discriminator loss & Generator loss \\ \hline
GAN \cite{Good14} & $-\underset{\mathbf{x}\sim p_{\text{data}}}{\mathds{E}} [\log{D(\mathbf{x})}]-\underset{\mathbf{z}\sim p_z}{\mathds{E}} [\log{(1-D(G(\mathbf{z})))}]$ & $\underset{\mathbf{z}\sim p_z}{\mathds{E}} [\log{(1-D(G(\mathbf{z})))}]$ \\
GAN-MA \cite{Good14} & $-\underset{\mathbf{x}\sim p_{\text{data}}}{\mathds{E}} [\log{D(\mathbf{x})}]-\underset{\mathbf{z}\sim p_z}{\mathds{E}} [\log{(1-D(G(\mathbf{z})))}]$ & $-\underset{\mathbf{z}\sim p_z}{\mathds{E}} [\log{D(G(\mathbf{z}))}]$\\
LSGAN \cite{Mao17} & $\underset{\mathbf{x}\sim p_{\text{data}}}{\mathds{E}} [(D(\mathbf{x})-1)^2]+\underset{\mathbf{z}\sim p_z}{\mathds{E}} [D(G(\mathbf{z}))^2]$ & $\underset{\mathbf{z}\sim p_z}{\mathds{E}} [(D(G(\mathbf{z}))-1)^2]$\\
WGAN \cite{Arjo17} & $-\underset{\mathbf{x}\sim p_{\text{data}}}{\mathds{E}} [D(\mathbf{x})]+\underset{\mathbf{z}\sim p_z}{\mathds{E}} [D(G(\mathbf{z}))]$ & $-\underset{\mathbf{z}\sim p_z}{\mathds{E}} [D(G(\mathbf{z}))]$\\
WGAN-GP \cite{Gulj17} & $-\underset{\mathbf{x}\sim p_{\text{data}}}{\mathds{E}} [D(\mathbf{x})]+\underset{\mathbf{z}\sim p_z}{\mathds{E}} [D(G(\mathbf{z}))] + \lambda \underset{\hat{\mathbf{x}}\sim p_{\hat{\mathbf{x}}}}{\mathds{E}} [(\Vert \nabla_{\hat{\mathbf{x}}}D(\hat{\mathbf{x}})\Vert _2 - 1)^2]$ & $-\underset{\mathbf{z}\sim p_z}{\mathds{E}} [D(G(\mathbf{z}))]$ \\
\end{tabular}}
\label{tab:objectives}
\end{table}

Nevertheless, GANs can still suffer from unstable training. This is partly due to the way in which the discriminator computes the difference between $p_{\text{data}}$ and $p_{\text{fake}}$. In the original GAN definition (Eq. \ref{eq:gan}), this difference is computed as the Jensen-Shannon divergence. This is a symmetric divergence measure, which unfortunately is poorly defined when two distributions are disjoint. To address this, alternative objective functions have been proposed that use the Pearson chi-square divergence \cite{Mao17} or the Earth Mover's or Wasserstein distance \cite{Arjo17}. Table \ref{tab:objectives} lists several alternative objective functions for the discriminator and generator that could be used to optimize a GAN. 

One attractive way to optimize GANs is to use the Wasserstein distance for the divergence between the real and fake distributions. An analogy for this distance is that of two piles that are different but both contain the same amount of earth: the Wasserstein distance is the minimum amount of work required to transform one pile into the other, in which work is defined as the amount of earth moved multiplied with the distance it is moved. In GANs, the discriminator could use this metric to compute the divergence between $p_{\text{data}}$ and $p_{\text{fake}}$. Computing the Wasserstein distance is intractable, but through the Kantorovich-Rubinstein duality a Wasserstein GAN objective can be formulated \cite{Arjo17} as

\begin{equation}
\underset{G}{\min}~\underset{D \in \mathcal{D}}{\max}~ V(D,G)=\underset{\mathbf{x}\sim p_{data}}{\mathds{E}} [D(\mathbf{x})]-\underset{\mathbf{z}\sim p_z}{\mathds{E}} [D(G(\mathbf{z}))],
\label{eq:earthmover}
\end{equation}

where $D$ is in $\mathcal{D}$, the set of 1-Lipschitz continuous functions for which the norm of the gradient should not exceed 1 \cite{Arjo17}. While standard GANs are driven by a classifier separating real from fake samples, Wasserstein GANs are driven by a distance measure that quantifies the similarity of two distributions Hence, discriminators in Wasserstein GANs do not return a probability but a scalar value, and are also referred to as \textit{critics}.  While the value of Eq. \ref{eq:gan} does not necessarily correspond with image quality, the distance in Eq. \ref{eq:earthmover} has been empirically shown to correlate with image quality \cite{Arjo17}.

There are several ways to obtain 1-Lipschitz continuity in the discriminator, among which weight clipping \cite{Arjo17} and the use of a gradient penalty \cite{Gulj17} are most commonly used. 
With weight clipping, the weights of the discriminator network are clipped to e.g. $[-0.01,0.01]$ at the end of each iteration. With gradient penalty, linearly interpolated samples are obtained between random real and synthesized samples. For each of these samples, the discriminator gradient should be less than 1. The objective function then becomes

\begin{equation}
\underset{G}{\min}~\underset{D \in \mathcal{D}}{\max}~ V(D,G)=\underset{\mathbf{x}\sim p_{data}}{\mathds{E}} [D(\mathbf{x})]-\underset{\mathbf{z}\sim p_z}{\mathds{E}} [D(G(\mathbf{z}))] - \lambda \underset{\hat{\mathbf{x}}\sim p_{\hat{\mathbf{x}}}}{\mathds{E}} [(\Vert \nabla_{\hat{\mathbf{x}}}D(\hat{\mathbf{x}})\Vert _2 - 1)^2],
\label{eq:gradientpenalty}
\end{equation}

where $p_{\hat{\mathbf{x}}}$ is the distribution of points along straight lines between randomly selected pairs of samples in $p_{\text{data}}$ and $p_{\text{fake}}$. The gradient penalty is weighted by a factor $\lambda$.

Another commonly encountered phenomenon is that of \textit{mode collapse}, in which the generator consistently synthesizes similar samples with little diversity. While such samples may be able to fool the discriminator, for obvious reasons it would be preferable if the generator provides more diverse samples. 
Mode collapse is addressed in Salimans et al. \cite{Sali16} by letting the discriminator not only look at individual samples, but also at the variation within a mini-batch. Another solution is to unroll the GAN \cite{Metz16} by computing generator updates through multiple versions of the discriminator. Furthermore, Wasserstein GANs have been shown to generate more diverse samples, potentially reducing the problem of mode collapse \cite{Arjo17}.

\subsection{The latent space}
During training of a GAN, the generator $G$ learns to map points in the low-dimensional latent space to points in the high-dimensional sample space. The latent space $p_z$ consists of a distribution with dimensionality $m$, such as a spherical Gaussian. It has been shown empirically that points that are close to each other in $p_z$ tend to result in samples that are also close in sample space \cite{Radf16}. 
Therefore, a walk through the latent space of a GAN may result in smooth interpolations between samples. This could also allow arithmetic in the latent space. Radford et al. \cite{Radf16} found that for a GAN trained to synthesize face images, subtracting the latent space point for 'man without glasses' from that of 'man with glasses' and adding that to 'woman without glasses' could result in an image of a woman with glasses.  

Although neighboring points in the latent space may correspond to samples with similar characteristics, there is no guarantee that the individual dimensions in the latent space correspond to interpretable and meaningful features. In fact, dimensions in the latent space may be highly entangled. One way to disentangle these dimensions is to use an InfoGAN \cite{Chen16}. InfoGANs include a latent code $\mathbf{c}$ in addition to the noise vector $\mathbf{z}$, so that $\mathbf{x} = G(\mathbf{z}, \mathbf{c})$. The GAN is trained so that the latent code represents disentangled features, such as the the angle or thickness of a hand-written digit. To prevent the generator from simply ignoring the latent codes, the mutual information $I(\mathbf{c};G(\mathbf{z}, \mathbf{c}))$ between $\mathbf{c}$ and $G(\mathbf{z}, \mathbf{c})$ is added to the objective in Eq \ref{eq:gan}. This is implemented using a separate neural network that tries to retrieve the latent code $\mathbf{x}$ from generated samples $G(\mathbf{z}, \mathbf{c})$. 

It may also be useful to not only obtain a mapping from the latent space to the sample space, but also the reverse mapping from the sample space to the latent space. This will lead to $m$-dimensional feature descriptors that have been learned in an unsupervised manner. Such feature descriptors can characterize real data samples and may be used in subsequent analysis. One way to obtain a mapping to latent space is to consider finding the location $\mathbf{z} \in p_z$ of a sample as a separate optimization problem. Given a sample $\mathbf{x}\in p_{\text{data}}$, the point in $\mathbf{z}\in p_z$ should be retrieved that minimizes the difference between $G(\mathbf{z})$ and $\mathbf{x}$, with fixed $G$ \cite{Lipt17}. Alternatively, GANs can be extended with an additional encoder neural network that maps real and synthesized samples to a latent space representation \cite{Dona16}.

\begin{figure}[tp]
\centering
\includegraphics[width=0.8\textwidth]{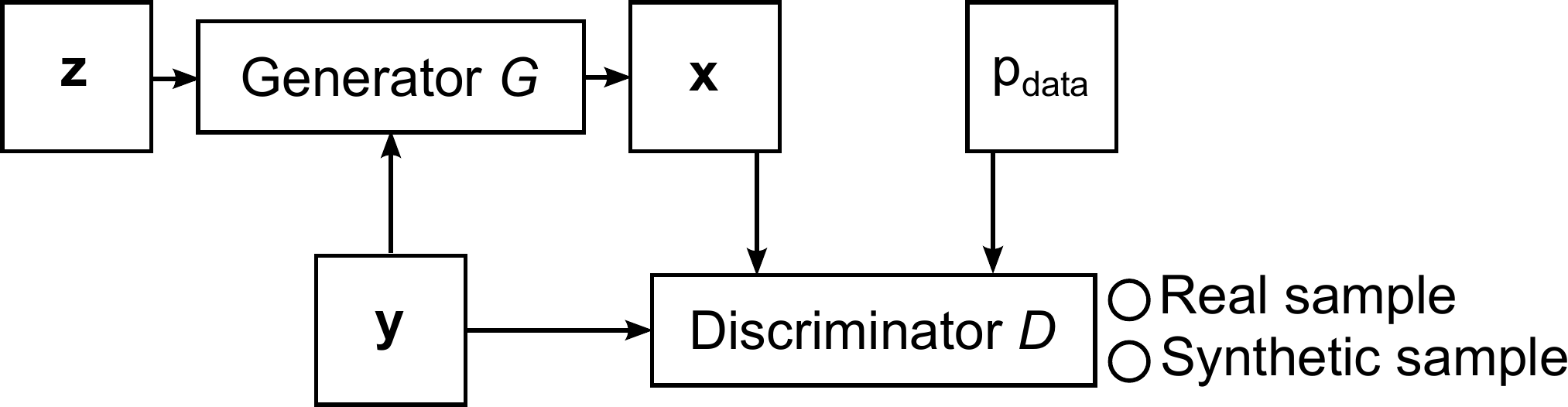}
\caption{In a conditional GAN, the generator and discriminator both take an additional conditioning vector $\mathbf{y}$ as input.}
\label{fig:cGAN}
\end{figure}

\subsection{Conditional GANs}
GAN training results in a generator model that can synthesize samples, but generally does not allow control over the characteristics of samples that are being generated. 
In practice, it may be beneficial to have more control over what is represented in the generated samples. In conditional GANs (cGANs), both the generator and the discriminator can take additional side-information into account \cite{Mirz14} (Fig. \ref{fig:cGAN}). This side-information is represented by an input vector $\mathbf{y}$ so that the objective function becomes

\begin{equation}
\underset{G}{\text{min}}~\underset{D}{\text{max}}~V(D,G)=\underset{\mathbf{x}\sim p_{\text{data}}}{\mathds{E}} [\log{D(\mathbf{x}|\mathbf{y})}]+\underset{\mathbf{z}\sim p_z}{\mathds{E}} [\log{(1-D(G(\mathbf{z}|\mathbf{y})|\mathbf{y}))}].
\label{eq:cGAN}
\end{equation}

The generator uses the information in $\mathbf{y}$ in addition to the noise vector $\mathbf{z}$ to synthesize plausible samples. The discriminator assesses whether these samples resemble real samples in $p_{\text{data}}$, given the information provided in $\mathbf{y}$. An example application is the synthesis of specific hand-written MNIST digits \cite{Mirz14}. The conditioning vector $\mathbf{y}$ in that cases contains a one-hot encoding of the ten digit classes, i.e. a vector in which one element is set to 1 and all other elements are set to 0. However, the conditioning vector is not restricted to such encodings. Mirza et al. \cite{Mirz14} also show how a text tag generator can be conditioned on a feature vector describing a natural image. Conversely, Zhang et al. \cite{Zhan17b} used cGANs to synthesize images from text descriptions. 

\subsection{GAN architectures}
Most GAN research has focused on the synthesis of 2D natural images using neural networks. Early applications of GANs used multi-layer perceptrons (MLPs) for the discriminator and generator. While this may be sufficient for the synthesis of smaller images, the number of network parameters may rapidly increase when MLPs are used for larger images. Therefore, a CNN is typically used for the generator as well as for the discriminator. To allow the generator to gradually increase the size of its representations, it is common to use fractionally strided or transposed convolutions for up-sampling. Denton et al. \cite{Dent15} proposed a Laplacian GAN to gradually increase the size of synthesized images using a sequence of conditional GANs, in which each image is conditioned on the upsampled version of a previous lower-resolution image. This allows synthesis of $64\times 64$ pixel RGB images, but requires training and evaluation of several GANs. Alternatively, Radford et al. \cite{Radf16} proposed three architectural choices to allow direct training of deep convolutional GANs (DCGANs). First, downsampling and upsampling should be performed with strided convolutions instead of pooling operations. Second, the use of fully connected layers should be prevented. Third, batch normalization should be used to normalize the inputs to activation functions. This allowed direct synthesis of $64\times 64$ pixel RGB images.

Due to the advent of alternative loss functions such as those based on the Wasserstein distance, GAN training stability has much improved in recent years. Nevertheless, synthesis of large images is still challenging. The current state-of-the-art allows synthesis of $1024 \times 1024$ pixel RGB images, by progressively blending in additional up-sampling layers (generator) and down-sampling layers (discriminator) \cite{Karr17}. By carefully increasing the image resolution, collapse of the training process can be prevented.  

\section{Adversarial methods for image domain translation}
\label{sec:advmethods}
In a conditional GAN, the generator is trained to synthesize plausible samples given a noise vector $\mathbf{z}$ and additional information provided in $\mathbf{y}$. Depending on the information encoded in the conditioning vector $\mathbf{y}$, the outputs of the generator could become heavily constrained: for a given input $\mathbf{y}$ the generator will always predict the same output, regardless of $\mathbf{z}$. For example, Isola et al. \cite{Isol17} found that when training a conditional GAN to translate an image from one domain to another domain, the generator mostly ignored the noise vector. Nevertheless, the adversarial network can provide valuable feedback to networks performing image domain translation: adversarial feedback can replace hand-crafted loss functions to quantify to what extent an image belongs to a particular target domain.

Two scenarios for training of an image domain translation model can be distinguished. In the first situation, a reference image in domain $B$ may be available for each image in domain $A$. For example, when domain $A$ contains grayscale photos and domain $B$ contains color photos. Hence, these problems can be approached by training \textit{with} paired images. In the second class of problems, one wishes to translate images from domain $A$ to a semantically related domain $B$, but domain $B$ does not necessarily contain a reference image for each image in domain $A$. This may be the case when translating between photos of horses and photos of zebras. In this scenario, training is performed \textit{without} paired training images.

\begin{figure}[tp]
\centering
\includegraphics[width=0.8\textwidth]{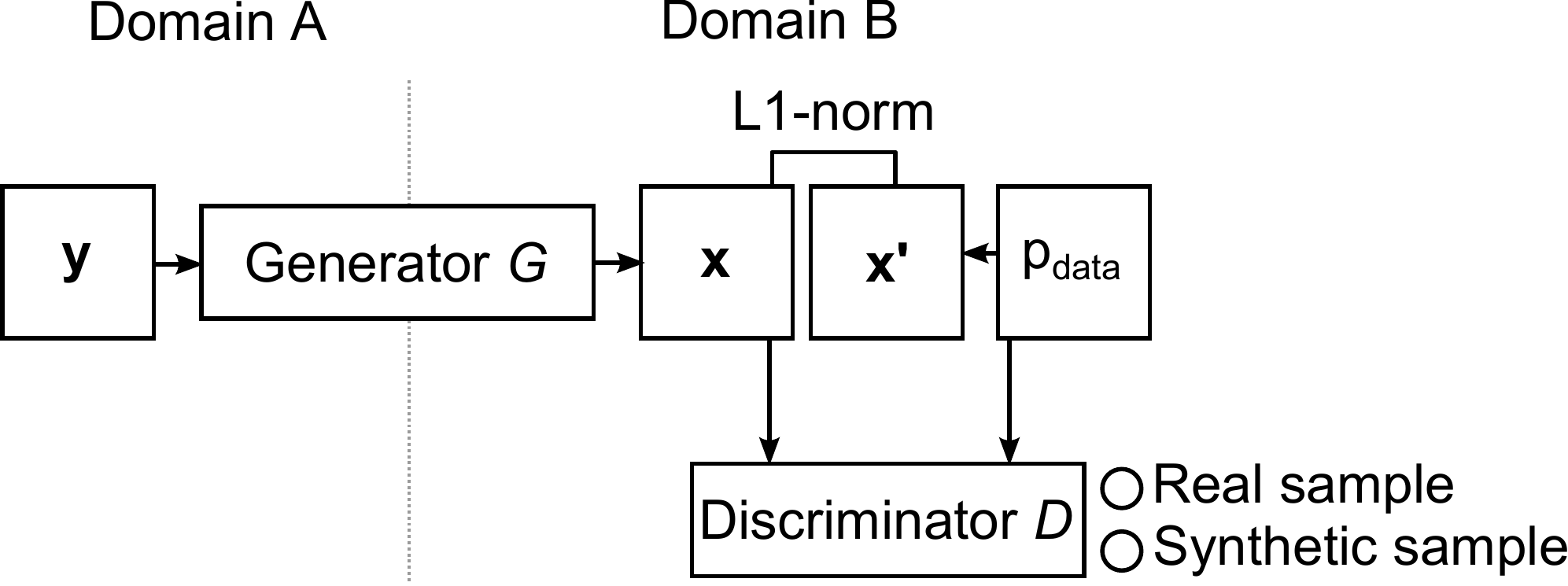}
\caption{In image domain translation \textit{with} paired training images in domains A and B, the generator $G$ minimizes the voxel-wise loss between its output $\mathbf{x}$ and a reference image $\mathbf{x'}$, and the discriminator $D$ provides an adversarial loss that reflects how well $\mathbf{x}$ corresponds to real images in domain $B$.}
\label{fig:pix2pix}
\end{figure}

\subsection{Training with paired images}
\label{sec:pix2pix}
Isola et al. \cite{Isol17} trained a generator to translate images from one domain to another domain. For each input image $\mathbf{y}$ in domain $A$, a reference image $\mathbf{x'}$ in domain $B$ was available (Fig. \ref{fig:pix2pix}). The generator was trained to minimize the pixel-wise difference between its prediction and this reference image. In addition, an adversarial network was used to assess whether the output images looked perceptually convincing compared to other images in domain $B$. The objective function used was

\begin{equation}
\begin{split}
\underset{G}{\text{min}}~\underset{D}{\text{max}}~V(D,G)=\lambda_1 \left(\underset{\mathbf{x}\sim p_{\text{data}}}{\mathds{E}} [\log{D(\mathbf{x}|\mathbf{y})}]\right. & +\left.\underset{\mathbf{y}\sim p_y}{\mathds{E}} [\log{(1-D(G(\mathbf{y})|\mathbf{y}))}]\right) \\ &+\lambda_2 \underset{\mathbf{x'}\sim p_{\text{data}},\mathbf{y}\sim p_y}{\mathds{E}}[\Vert \mathbf{x'} - G(\mathbf{y})||_1],
\end{split}
\label{eq:pix2pix}
\end{equation}

where $\mathbf{x'}$ and $\mathbf{y}$ are assumed to be spatially aligned (Fig. \ref{fig:pix2pix}). When $\lambda_1=0$ and $\lambda_2>0$, the system is effectively reduced to only the generator, which tries to minimize the pixel-wise unstructured loss between its output $G(\mathbf{y})$ and a reference image. When $\lambda_1>0$ and $\lambda_2=0$, network $G$ is trained to transform the input into an image that looks similar to other images in domain $B$. However, because this output image is not directly linked to a reference image in domain $B$, it does not necessarily correspond to the generator's input. The combination of $\lambda_1>0$ and $\lambda_2>0$ ties the output image $G(\mathbf{y})$ to the reference image $\mathbf{x}$ and the target domain through both the $L1$-norm and the adversarial loss.

Similar ideas were proposed for image segmentation \cite{Luc16} and image colorization. Ledig et al. \cite{Ledi16} proposed to use adversarial training for super-resolution and thus to translate low-resolution images into high-resolution images. The discriminator tries to distinguish the reconstructed high-resolution images from natural images in a reference database. Correspondence between the reconstructed image and the original high-resolution image is enforced using an $L2$ content loss term calculated in feature space (e.g. of a VGG-network) while the adversarial loss component encourages perceptually convincing samples.

Adversarial networks performing image domain translation tend to be more stable than regular GANs, as the feature maps are heavily conditioned on an existing image. Architectures used for the generator could be any fully convolutional network architecture, such as \cite{Long15,Ronn15,Yu15}. The discriminator could look either at the full image, or only at sub-images \cite{Isol17,Zhu17}. In the latter case, the discriminator could focus more on high-frequency information, while low-frequency information is captured in a different loss term such as the $L1$-norm.

\subsection{Training without paired images}
\label{sec:cyclegan}
The objective function in Eq. \ref{eq:pix2pix} assumes that for each input image $\mathbf{y}$ there is an aligned corresponding reference image $\mathbf{x'}$ in domain $B$. However, corresponding pairs of images in two different domains may often be unavailable in practice, e.g. when translating between different imaging modalities. Even if the same patient is scanned with two modalities, it may be impossible to correctly align the resulting images in order to use a voxel-wise loss term. If the assumption of aligned pairs of training images were to be removed from Eq. \ref{eq:pix2pix}, the objective function would be

\begin{equation}
\underset{G}{\text{min}}~\underset{D}{\text{max}}~V(D,G)=\underset{\mathbf{x}\sim p_{\text{data}}}{\mathds{E}} [\log{D(\mathbf{x})}] +\underset{\mathbf{y}\sim p_y}{\mathds{E}} [\log{(1-D(G(\mathbf{y})))}].
\label{eq:ucGAN}
\end{equation}

This objective resembles that of a GAN (Eq. \ref{eq:gan}) with deterministic inputs from input images $\mathbf{y}$. In this formulation the generator will learn to generate samples that mimic the target domain, but there is no guarantee that these samples match the actual content of the input image in the source domain. This problem can be mitigated by adding additional loss terms through either self-regularization or cyclic consistency.

\begin{figure}[tp]
\centering
\includegraphics[width=0.8\textwidth]{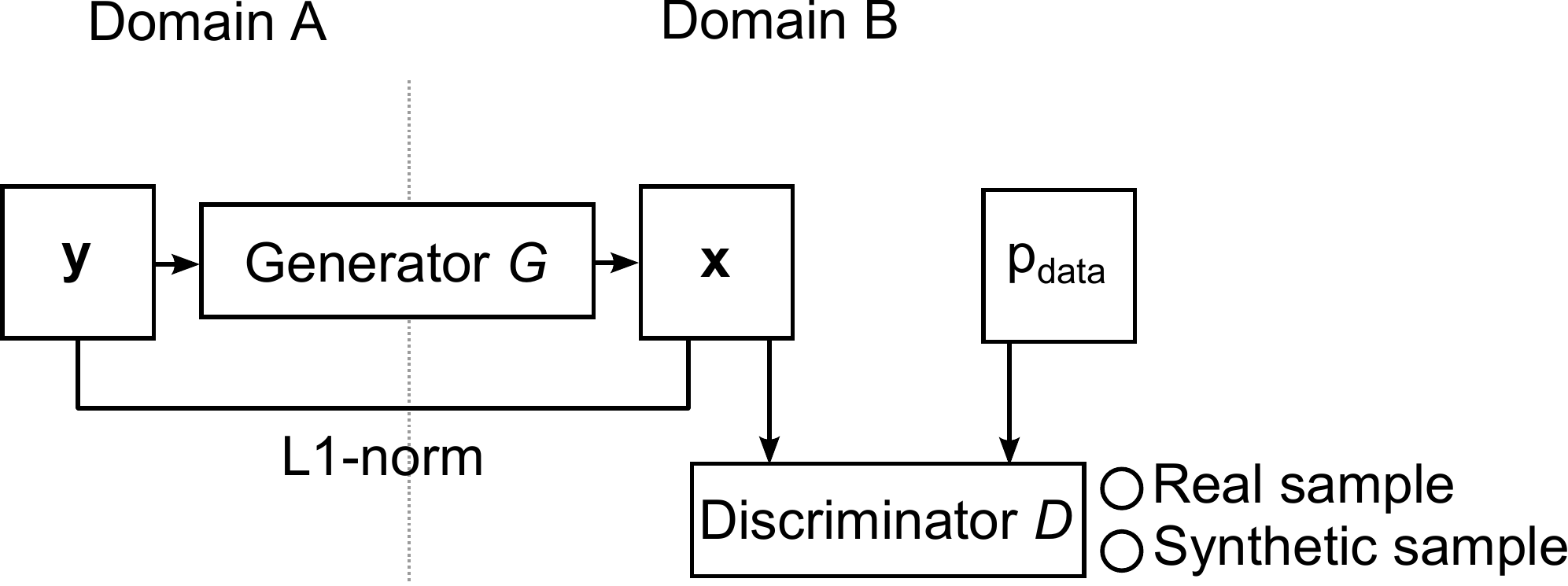}
\caption{In image domain translation \textit{without} paired training images in domains A and B, self-supervision could be used. The generator $G$ minimizes the voxel-wise loss between its output $\mathbf{x}$ and the input image $\mathbf{y}$. In addition, the discriminator $D$ provides an adversarial loss that reflects how well $\mathbf{x}$ corresponds to real images in domain $B$.}
\label{fig:selfreg}
\end{figure}

Self-regularization introduces an additional loss term that minimizes the difference between the input and the output image to encourage that they maintain the same content \cite{shrivastava2017learning} (Fig. \ref{fig:selfreg}). In that case, the objective becomes

\begin{equation}
\begin{split}
\underset{G}{\text{min}}~\underset{D}{\text{max}}~V(D,G)=\underset{\mathbf{x}\sim p_{\text{data}}}{\mathds{E}} [\log{D(\mathbf{x})}] &+\underset{\mathbf{y}\sim p_y}{\mathds{E}} [\log{(1-D(G(\mathbf{y})))}] \\&+\lambda \underset{\mathbf{y}\sim p_y}{\mathds{E}}[\Vert \mathbf{y} - G(\mathbf{y})||_1].
\end{split}
\label{eq:selfreg}
\end{equation}

However, this is only feasible when some similarity between the input and output image can be expected. For example, Shrivastava et al. \cite{shrivastava2017learning} use self-regularization to refine simulated images of the eye. In that case, the low-frequency information in the final image is expected to be similar to that in the input image. However, when $G(\mathbf{y})$ and $\mathbf{y}$ are in two very different domains, self-regularization may lead to undesirable results. 

\begin{figure}[tp]
\centering
\includegraphics[width=0.9\textwidth]{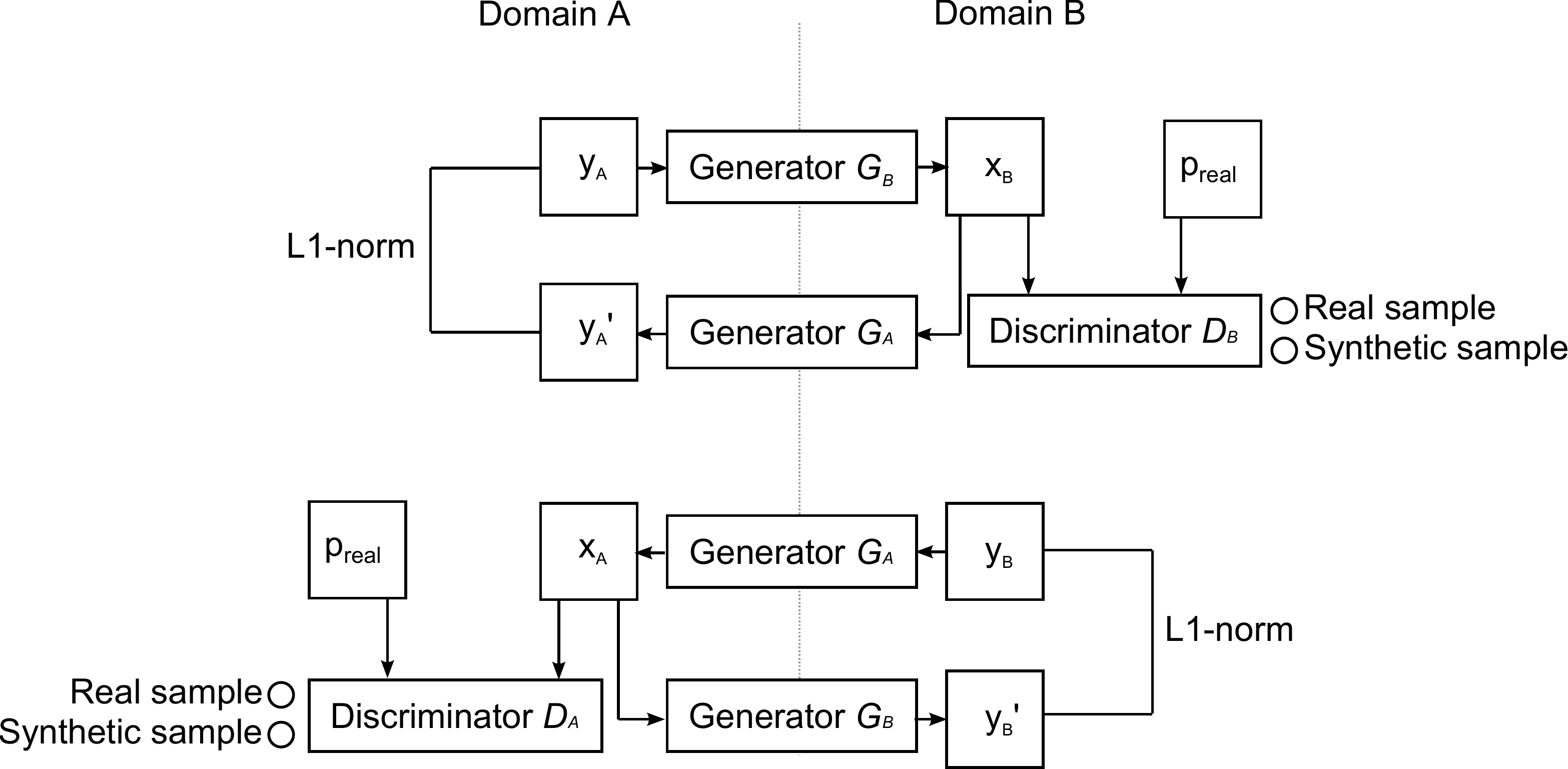}
\caption{A CycleGAN consists of two generator networks and two discriminator networks. The generators translate images from domain $A$ to domain $B$ and vice versa. The discriminators differentiate between real and synthetic samples in each domain. A pixel-wise $L1$-loss in each domain determines whether images are consistently recovered after cyclic translation.}
\label{fig:cyclegan}
\end{figure}

Cycle consistency assumes that an image $\mathbf{y}_A$ in domain $A$ that is translated into an image $\mathbf{x}_B$ in domain $B$ can also be mapped back to an image $\mathbf{y}_A'$ in domain $A$. The reconstructed image $\mathbf{y}_A'$ should be similar to the original image $\mathbf{y}_A$. In addition, as before, images that are mapped to a target domain should be indistinguishable from other images in the target domain (Fig. \ref{fig:cyclegan}). Models trained with cycle consistency therefore contain a generator $G_B$ that transforms images from domain $A$ to domain $B$, a generator $G_A$ that transforms images from domain $B$ to domain $A$, a discriminator $D_B$ in domain $B$ and a discriminator $D_A$ in domain $A$.  Two cycles are trained simultaneously, one from domain $A$ to domain $B$ and back, and one from domain $B$ to domain $A$ and back. This idea was proposed in \cite{Zhu17,Yi17,Kim17} and is commonly referred to as a CycleGAN. To train a CycleGAN, the generator networks try to minimize the discrepancy between input images as well as their reconstruction in the original domain. In addition, the generators try to maximize the loss of the discriminators, which in turn try to distinguish synthetic samples from real samples in their respective domains. Zhu et al. \cite{Zhu17} showed how a CycleGAN can be used to transform e.g. photographs into paintings with different styles.

\section{Domain adaptation via adversarial training}
\label{sec:theory_da}

\begin{figure}[tp]
\centering
\includegraphics[width=0.85\textwidth]{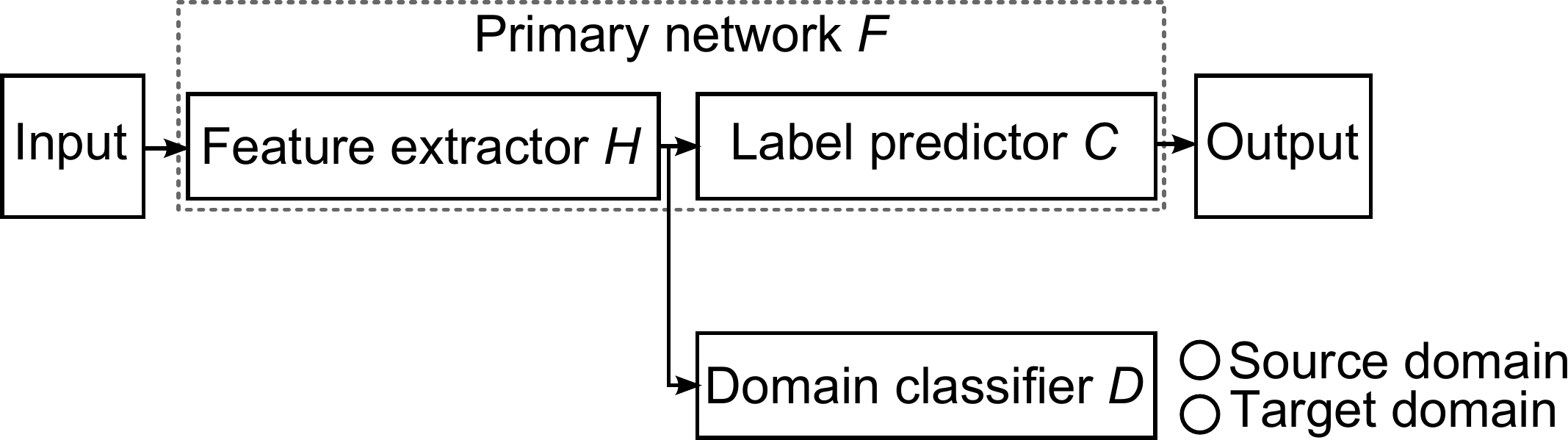}
\caption{Domain adaptation with adversarial methods. The primary network $F$ is split into a feature extractor $H$ and a label predictor $C$. The feature extractor tries to embed samples in a space that expresses only the domain-independent information that is necessary for $C$ to perform the primary task. An auxiliary domain classifier network $D$ tries to classify the domain of the input based on the latent representation. The feature extractor and domain classifier are trained as adversaries.}
\label{fig:ADA}
\end{figure}

A common underlying assumption in machine learning is that training and test data are drawn from the same distribution. A predictive model trained on data from a \emph{source} domain $S$ and distribution $p_S$ may under-perform when applied to data from a \emph{target} domain $T$ and a distribution $p_T$ that differs from $p_S$. Many factors can contribute to this problem of \emph{domain shift} in medical imaging. For example, clinical centers may be using different acquisition protocols or scanners, and thus acquire scans of different modalities, quality, resolution or contrast. These variations are a significant practical obstacle towards the use of machine learning in multi-center studies or its large-scale adoption outside the research labs. Generating a labeled database for every possible target domain is not a realistic solution.

Domain adaptation (DA) is the field that explores how to learn a predictor using labeled data in domain $S$ and adapt it to generalize on domain $T$ without any or with limited additional labels \cite{ben2010theory}. Of great interest is \emph{unsupervised} domain adaptation (UDA), which assumes no labeled data from domain $T$. The basic assumption of DA is that there exists a predictive function that performs well on data from both domains \cite{ben2010theory}. 
UDA methods learn mappings between domains or to a new representation $\mathcal{H}$, so that the predictor $C$ learned using labeled data from domain $S$ can be applied on data from domain $T$ when appropriately combined with the mappings.
Ben-David et al. \cite{ben2010theory} showed that to learn how to map samples to a representation $\mathcal{H}$ that allows predictor $C$ trained on embedded labeled samples from $S$ to generalize on embedded samples from $T$, the learning algorithm has to minimize a \emph{trade-off} between the source error and the divergence of the two distributions of the embedded samples. 
They pointed out that a strategy to minimize this divergence is to find a representation where samples from the two domains are indistinguishable by a domain classifier. This idea formed the basis for domain adaptation via adversarial training  \cite{ganin2016domain}, which treats the domain classifier as an adversarial network.

The basic framework for domain adaptation with adversarial networks \cite{ganin2016domain} is shown in Fig.~\ref{fig:ADA}.
The primary network $F$ (e.g. a classifier) is decomposed into the feature extractor $H$ (e.g. all hidden layers), which maps the input to a latent-representation $\mathcal{H}$, and the label predictor $C$ (e.g. the last linear classifier). An auxiliary network, the domain classifier $D$, tries to distinguish the domain of the input based on the extracted features.
To train $H$ to extract a domain invariant representation that simultaneously enables the primary task on both domains, two cost functions are considered.
The first cost, $\mathcal{L}_c$, is for learning the primary task, such as cross-entropy between predictions of classifier $F(\mathbf{x})=C(H(\mathbf{x}))$ and the training labels. The second cost is the negative log likelihood of the domain discriminator:
\begin{equation}
\mathcal{L}_d = - \underset{\mathbf{x}\sim p_S} {\mathds{E}} [\log{D(H(\mathbf{x}))}] - \underset{\mathbf{x}\sim p_T} {\mathds{E}} [\log{(1-D(H(\mathbf{x})))}]
\end{equation}
Note this is related to the objective of standard GANs (Table~\ref{tab:objectives}), where instead of real and synthetic data, it involves representations of samples from the two domains.
During training, $D$ is trained to classify the domain of input samples, hence to minimize $\mathcal{L}_d$. The primary network simultaneously learns the primary task on the source labelled data and minimizes $\mathcal{L}_c$, while learning a feature extractor $H$ such that the domain discriminator's loss $\mathcal{L}_d$ is maximized.
Such a training objective results in a mapping $H(\mathbf{x})$ that does not preserve domain-specific features, rather only the information necessary to perform the primary task. Assuming that both the domains contain the information necessary for the primary task, the network $F(\mathbf{x})=C(H(\mathbf{x}))$ is able to make predictions for unseen input regardless its domain. Otherwise a \emph{trade-off} between domain invariance and performance is necessary.

\section{Applications in biomedical image analysis}
\label{sec:GANSinMIA}
Recent years have seen applications of GANs and adversarial methods to a wide range of problems in biomedical image analysis. This section discusses a number of these applications. Sec. \ref{sec:biomedsynth} describes studies focused on \textit{de novo} generation of data through sampling from a latent distribution using GANs, and the use of GANs to detect novelties or abnormalities in medical images. In Sec. \ref{sec:biomedconvert}, adversarial methods that convert images from one modality to another are described. Furthermore, adversarial methods can be used to improve the quality of biomedical images, e.g. by reducing image acquisition artifacts or improving image resolution (Sec. \ref{sec:biomedquality}). Sec. \ref{sec:biomedtasks} describes the use of adversarial methods for biomedical image segmentation. Sec. \ref{sec:app_dom_adapt} describes adversarial domain adaptation methods for transfer learning between different kinds of images. Sec. \ref{sec:semisuper} describes how GANs can be integrated in methods for semi-supervised learning.

A common factor in many applications is that the discriminator is only used during training. The generator performs the main task, and is used during training as well as during testing. One exception to this is semi-supervised learning (Sec. \ref{sec:semisuper}), in which the generator is discarded and the discriminator is kept.

\subsection{Sample generation}
\label{sec:biomedsynth}
GANs have been used for \textit{de novo} generation of samples of medical images or anatomical structures from a latent distribution $p_z$. Generated samples could potentially be used to enlarge training sets for discriminative models, or synthesize data for training of human experts. This is a challenging task: whereas there may be a certain tolerance for errors in synthesized samples in some domains such as natural images, such errors can have strong negative effects in medical imaging. 

An example of de novo synthesis of images is the work by Chuquicusma et al. \cite{Chuq17}, who used a DCGAN to synthesize $56\times 56$ pixel 2D CT image patches showing lung nodules. The appearance of synthesized lung nodules was assessed in an observer study, which showed that in many cases the synthesized patches were able to fool the radiologist. Nevertheless, the small size of these images may be a limiting factor for some applications. To allow synthesis of larger images, Beers et al. \cite{Beer18} used the progressive GAN method by Karras et al. \cite{Karr17} to generate $256\times 256$ pixel slices of multi-modal MRI images and $512\times 512$ pixel retinal fundus images. The method jointly synthesized the medical image and a corresponding segmentation mask for brain tumors in MRI, and for retinal vessels in fundus images. The results showed that joint synthesis of the image and the segmentation mask led to higher quality synthesized images. In the work by Galbusera et al. \cite{Galb18} a GAN is used to synthesize a full sagittal 2D X-ray image of the lumbar spine based on a simple sketch of the vertebrae. 

Synthesis of 3D anatomical structures using GANs is challenging, as de novo generation of voxelized 3D volumes can lead to holes and fragments in synthesized structures. This issue may be addressed by using specific data representations. Wolterink et al. \cite{Wolt18} used a Wasserstein GAN with gradient penalty to obtain 1D representations of contiguous 3D coronary artery geometries. Using  a training set of semi-automatically extracted coronary centerlines from coronary CT angiography (CCTA) scans, a GAN was trained to synthesize vessels. An analysis of the latent space showed that latent feature representations can distinguish between left and right coronary arteries and short and long arteries.

One attractive feature of GANs is their ability to learn the probability distribution of a data set. This can be used to identify cases deviating from a given distribution, i.e. abnormalities or novelties. This is a highly relevant topic in medical image analysis, where a common goal is the detection of deviations from the normal. Schlegl et al. \cite{Schl17} used GANs for the analysis of retinal OCT images. A GAN was trained to generate images of healthy patients based on noise vectors sampled from a uniform distribution. To identify anomalies in new and unseen images, an iterative process was used to find the noise vector in the latent space corresponding to the image with the lowest reconstruction error with respect to the input sample. Each image sample was assigned an anomaly score, which was based on the reconstruction error, as well as on the output of the discriminator for the sample. Hence, both the generator and discriminator were used to identify anomalies.

\subsection{Image synthesis}
\label{sec:biomedconvert}
In clinical practice, information from multiple medical imaging modalities is often combined. However, depending on the application, the information contained in a particular image modality may already be present in other modalities. Accurate conversion of images from one imaging domain to another imaging domain could help lower the number of acquisitions required, thereby reducing patient discomfort and costs.

One example of a clinical problem in which accurate conversion from one modality to another modality may be desirable is radiotherapy treatment planning. In conventional treatment planning, tumors and organs-at-risk are delineated in MR, and electron-density values for radiation dose calculation are measured using CT. These two are then combined in radiation treatment planning. In MR-only radiotherapy treatment planning, a CT image is synthesized based on the MR image. This allows the CT image to be skipped during treatment planning, thereby reducing imaging burden for the patient. Several methods have used regression CNNs to learn a direct mapping between MR images and CT images \cite{Han17,Dink18} by minimizing a voxel-wise loss between synthesized and corresponding paired target images. However, a potential side-effect of voxel-wise loss minimization is blurring: The MR and CT image are two separate acquisitions and there is no perfect spatial overlap between the images, even after image registration. Consequently, similar MR inputs may correspond to different CT outputs in the training set, and during testing the CNN will predict a blurred image.

\begin{figure}[tp]
\centering
\includegraphics[width=0.95\textwidth]{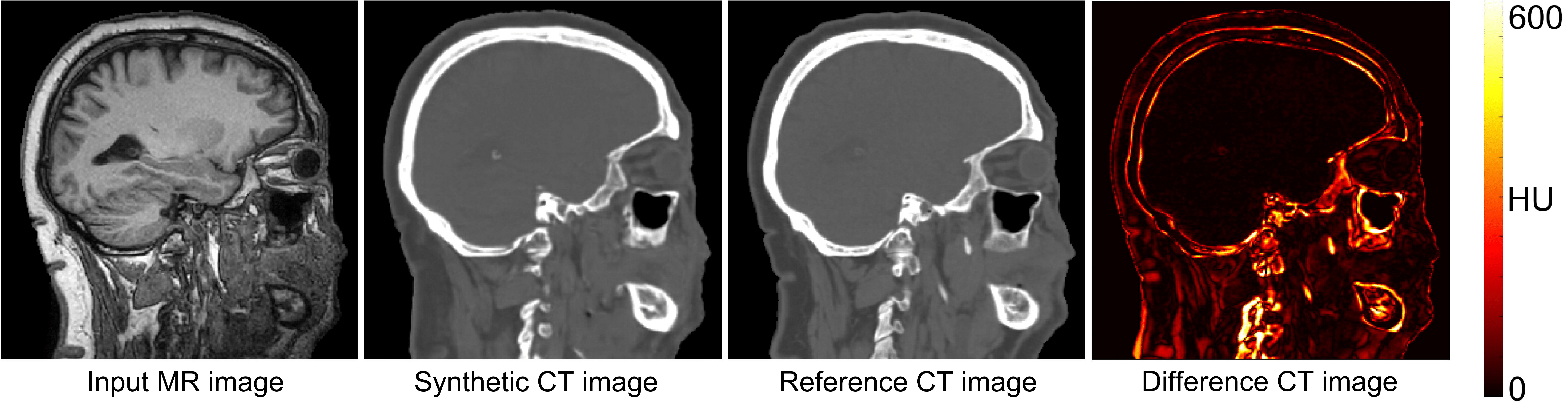}
\caption{Result from Wolterink et al. \cite{Wolt17b}, showing how a GAN with cycle consistency can be used to convert T1-weighted MR images into CT images. This example shows the input MR image, the synthetic CT image, the reference CT image and the difference between the synthetic and reference CT image.}
\label{fig:cycleganresults}
\end{figure}

To ensure that synthesized CT images better resemble real CT images, Nie et al. \cite{Nie17} combined the voxel-wise loss of a regression CNN with adversarial feedback from a discriminator, similarly to the method described in Sec. \ref{sec:pix2pix}. However, this method still requires aligned images in the MR and the CT domains. As mentioned above, accurate alignment of these two modalities is challenging. Furthermore, aligned training images may not always be available. To overcome this, Wolterink et al. \cite{Wolt17b} used a CycleGAN as described in Sec. \ref{sec:cyclegan} to allow MR to CT synthesis without paired MR and CT training images. A forward CycleGAN was trained to transform MR into CT and back to MR, and a backward CycleGAN was trained to transform CT images into MR and back to CT. Fig. \ref{fig:cycleganresults} shows an example of results obtained in \cite{Wolt17b}. 

Similarly, Chartsias et al. \cite{Char17} trained a CycleGAN to translate cardiac MR images into cardiac CT images and vice versa. This highlights the advantage of unpaired training: it is challenging to align cardiac MR and cardiac CT images for training of an image conversion CNN could be trained with a voxel-wise regression loss. Chartsias et al. used this image conversion for domain adaptation. Reference segmentations in domain $A$ could be directly transferred to the synthesized images in domain $B$ to enlarge the amount of available training data for a segmentation CNN in domain $B$. Likewise, Huo et al. \cite{huo2018adversarial} showed how a CycleGAN can train a segmentation network without any segmentations in the target domain. GANs have also been used to convert between other modalities, such a amyloid PET and structural MR images \cite{Choi17} for the analysis in patients with Alzheimer's disease.

\begin{figure}
\centering
\subfloat[Low-dose]{
\includegraphics[width=0.22\columnwidth]{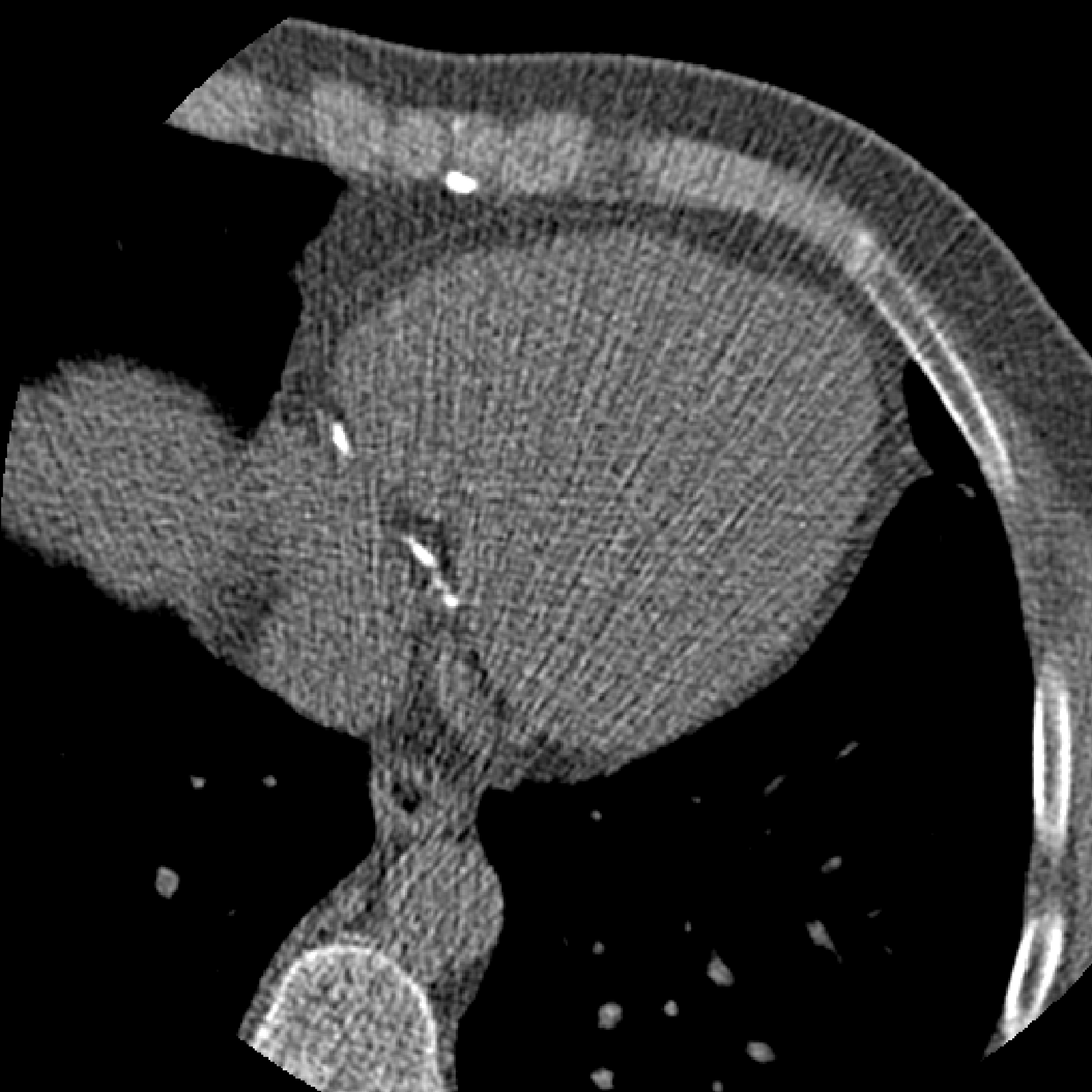}
\label{subfig:vivo20fbp}
} 
\subfloat[Adversarial model]{
\includegraphics[width=0.22\columnwidth]{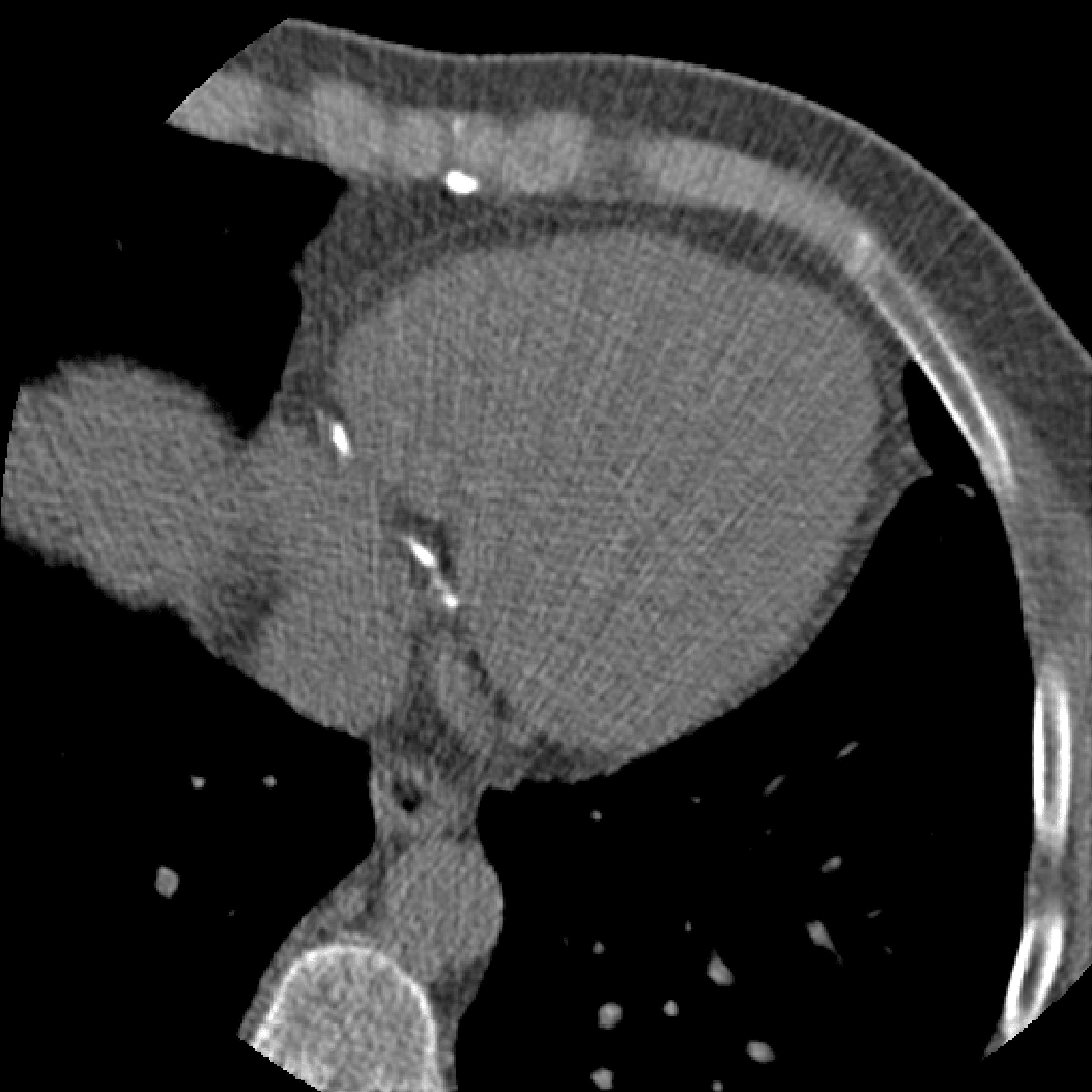}
\label{subfig:vivo20g3}
} 
\subfloat[IR]{
\includegraphics[width=0.22\columnwidth]{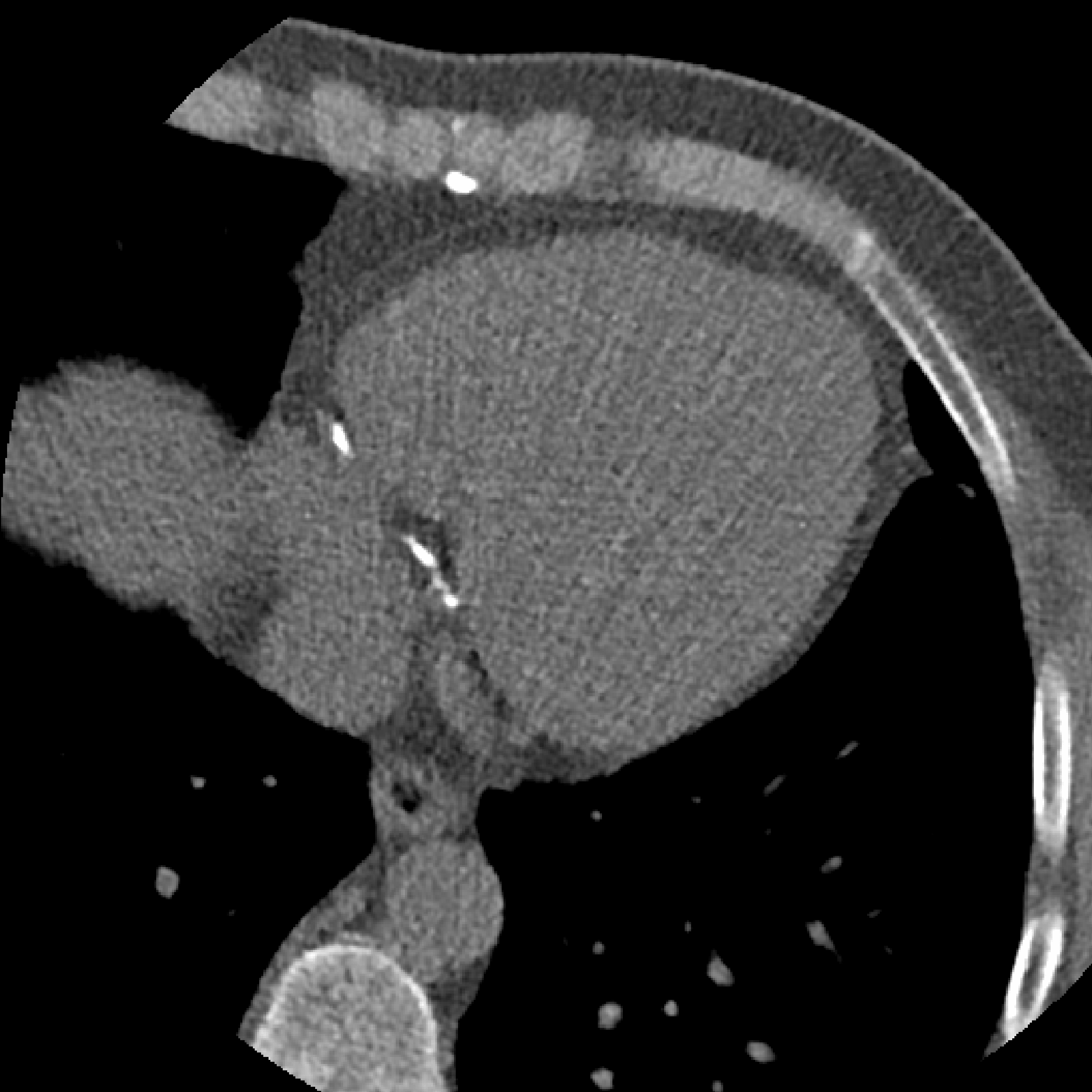}
\label{subfig:vivo20ir}
}
\subfloat[Routine-dose]{
\includegraphics[width=0.22\columnwidth]{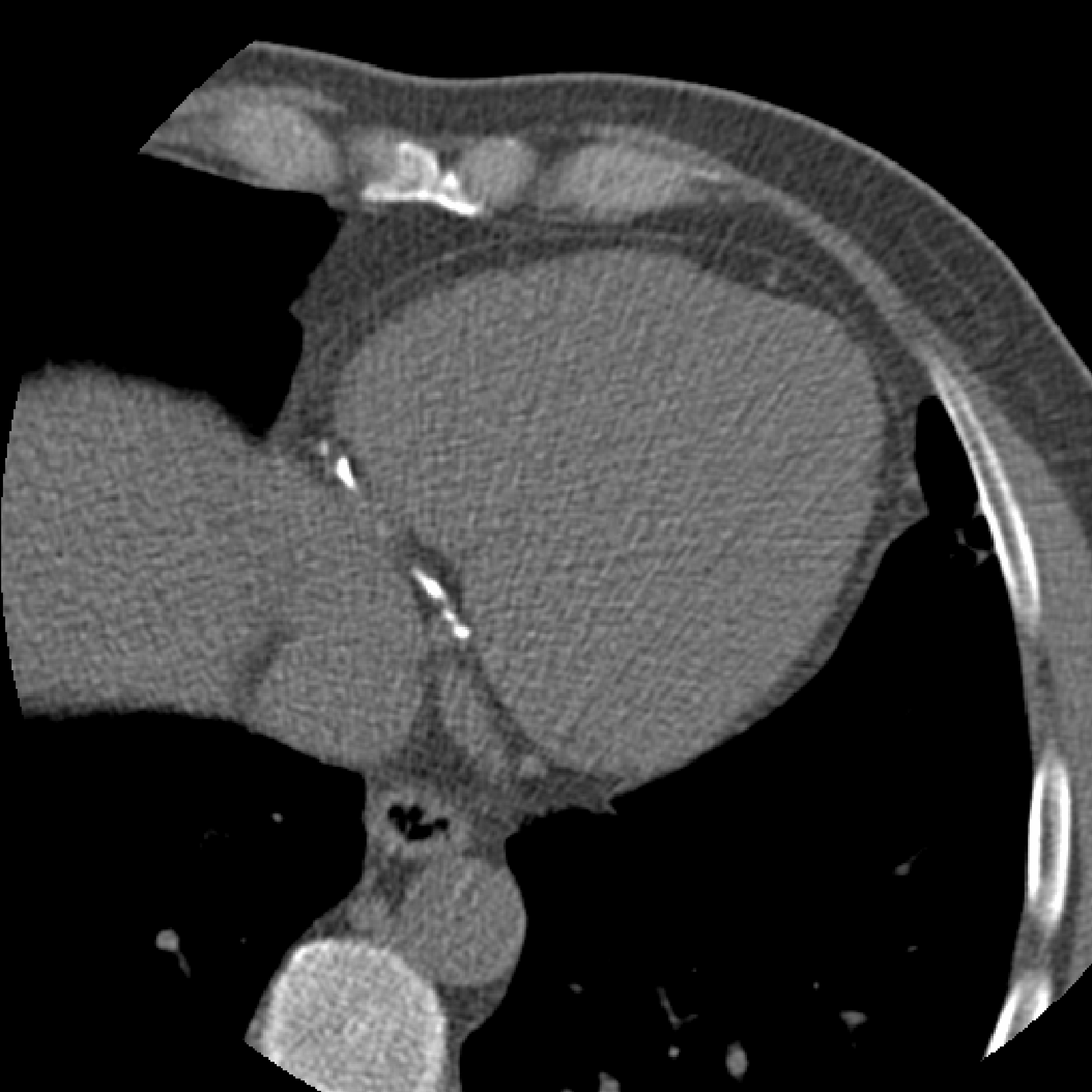}
\label{subfig:vivo100fbp}
} 
\\
\subfloat[CAC mask]{
\includegraphics[width=0.22\columnwidth]{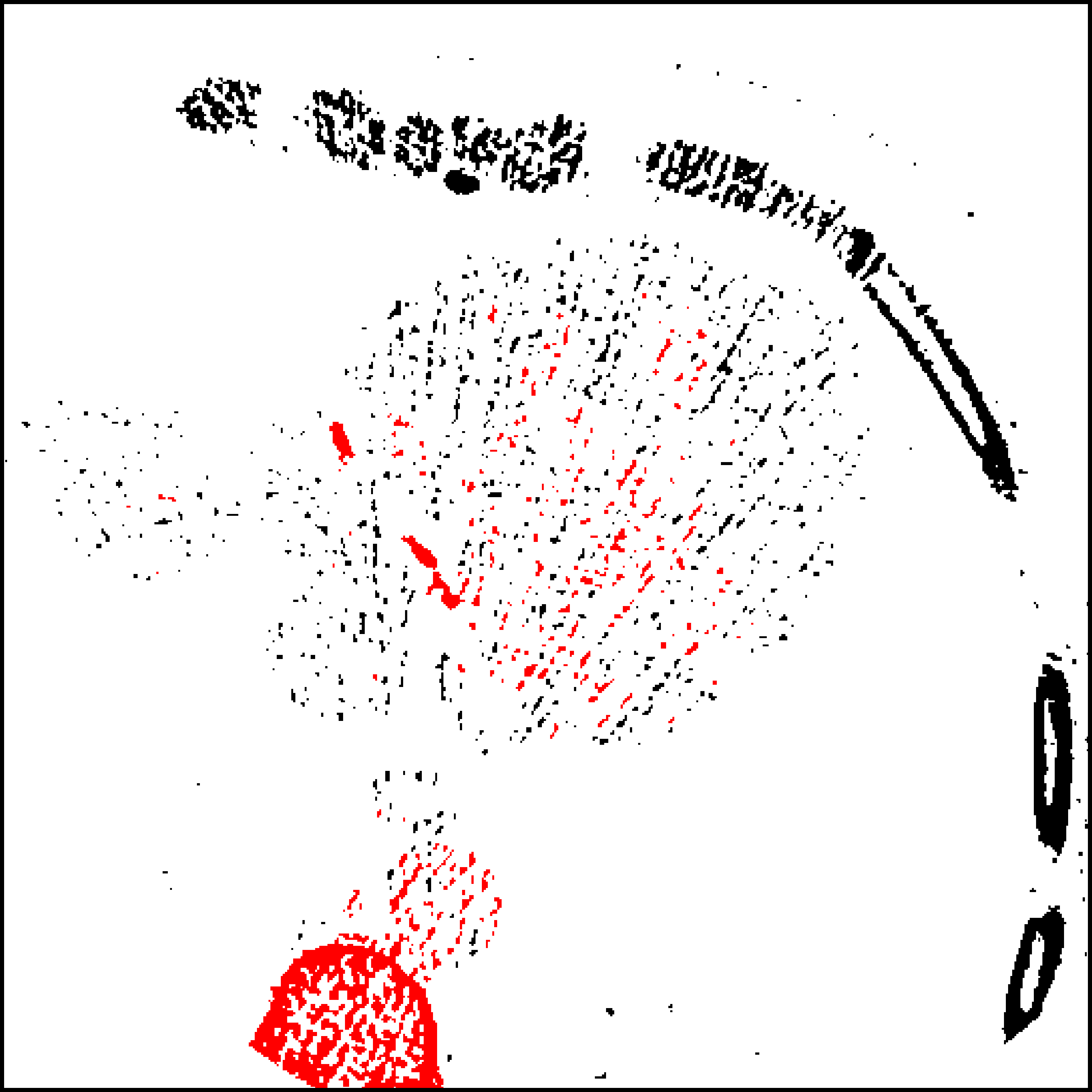}
\label{subfig:vivo20fbpcac}
} 
\subfloat[CAC mask]{
\includegraphics[width=0.22\columnwidth]{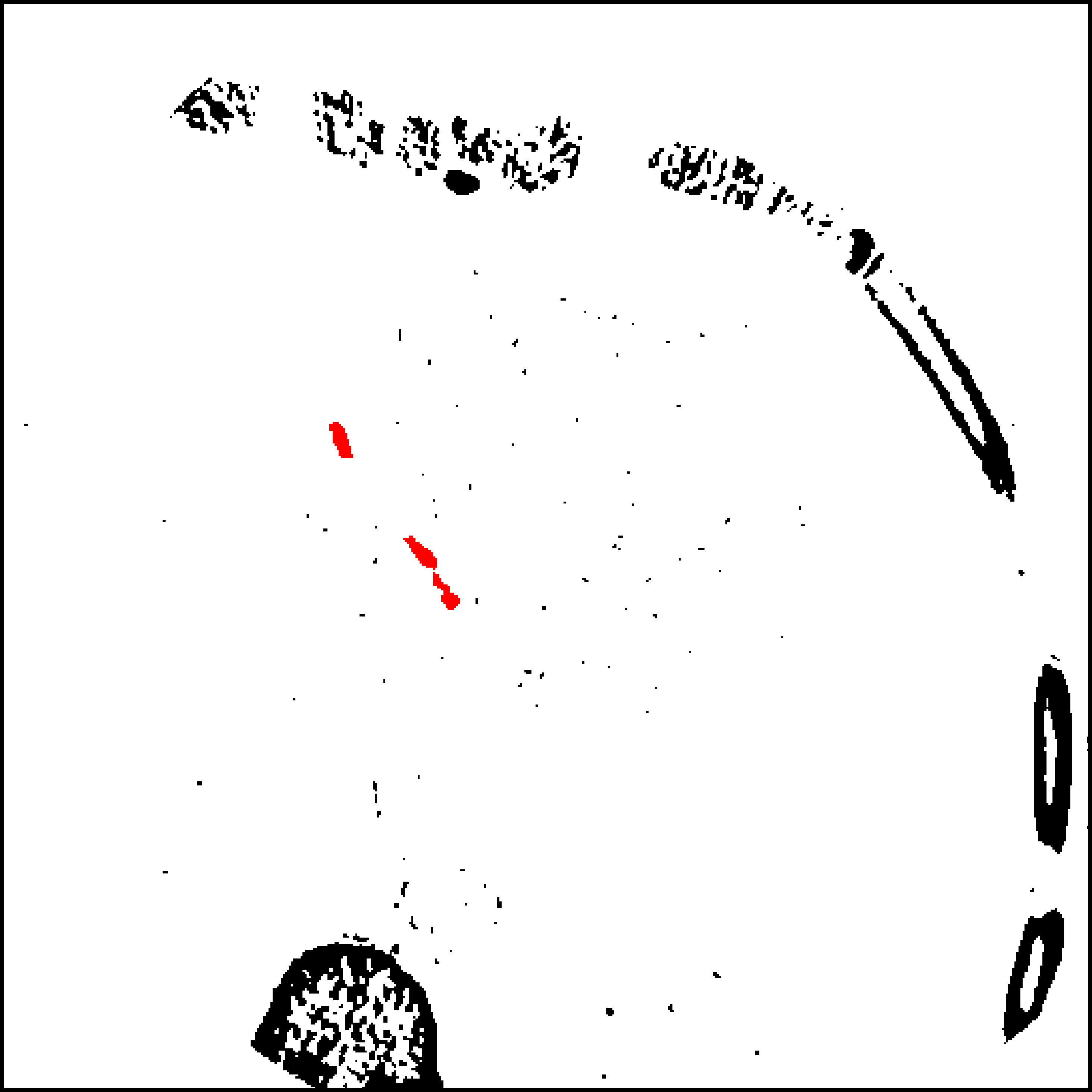}
\label{subfig:vivo20g3cac}
} 
\subfloat[CAC mask]{
\includegraphics[width=0.22\columnwidth]{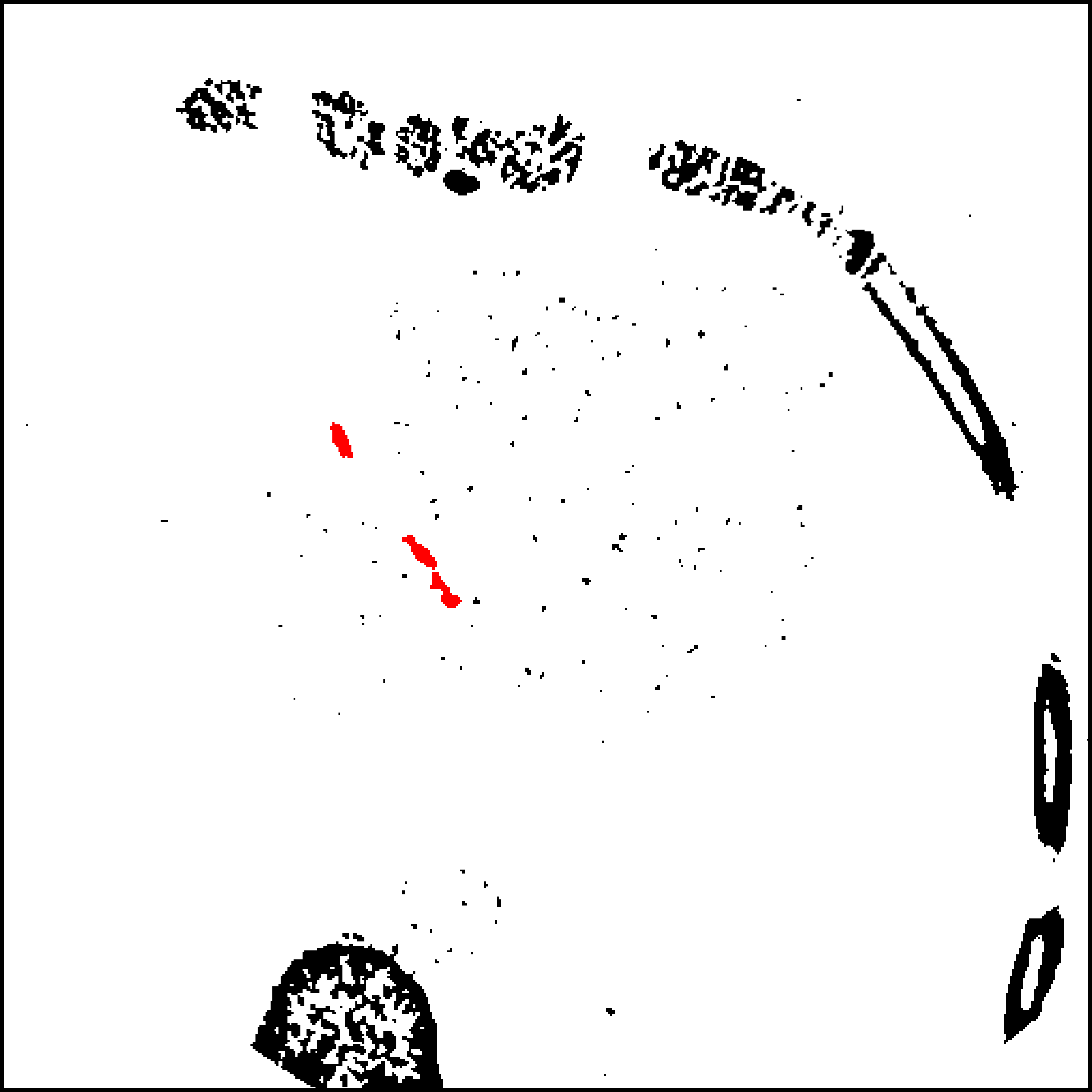}
\label{subfig:vivo20ircac}
} 
\subfloat[CAC mask]{
\includegraphics[width=0.22\columnwidth]{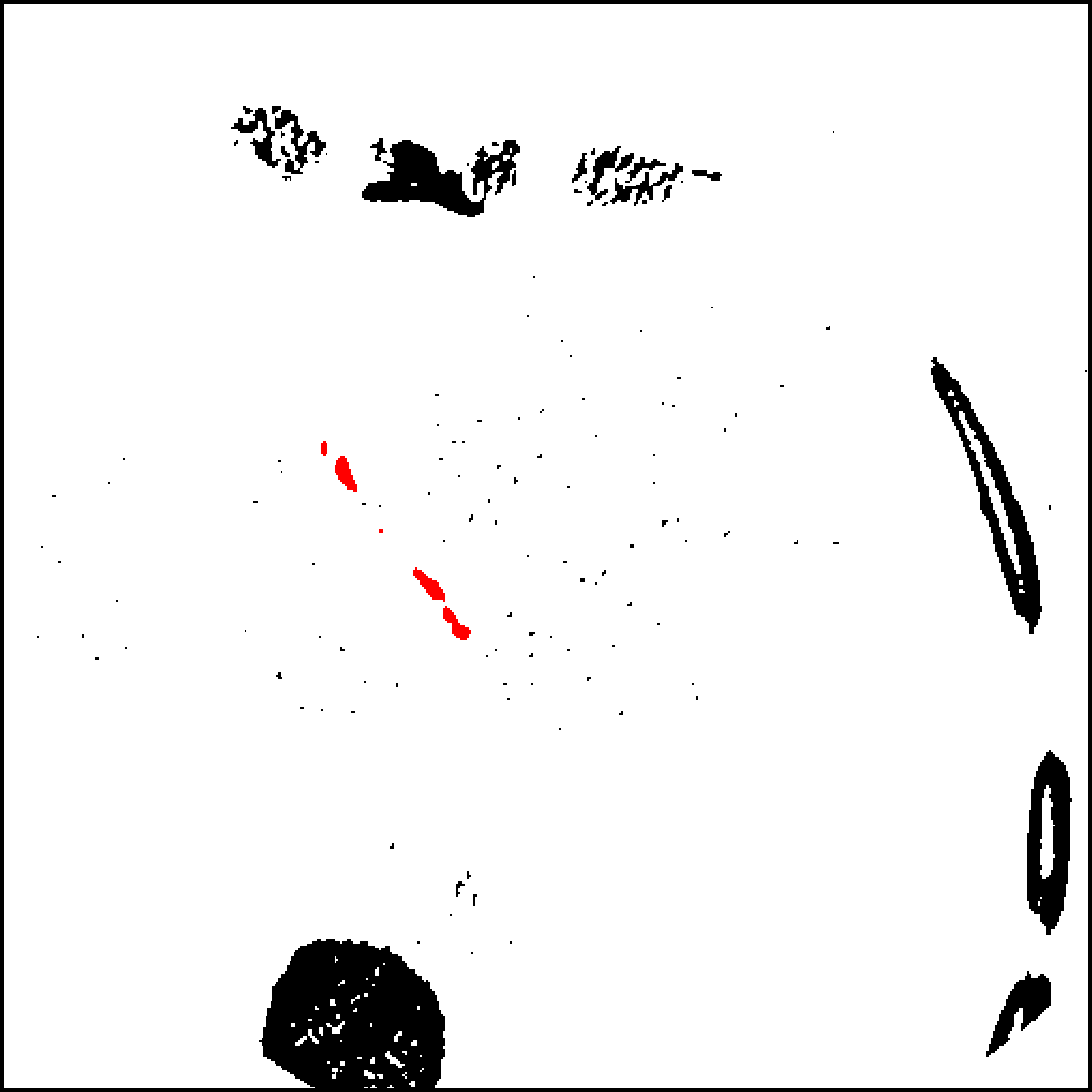}
\label{subfig:vivo100fbpcac}
} 
\caption{Result from Wolterink et al. \cite{Wolt17}. Example CT slice of \protect\subref{subfig:vivo20fbp} 20\% low-dose FBP reconstruction and \protect\subref{subfig:vivo20fbpcac} corresponding coronary artery calcification (CAC) scoring mask, \protect\subref{subfig:vivo20g3} 20\% dose GAN-based noise reduction and \protect\subref{subfig:vivo20ircac} corresponding CAC scoring mask, 20\% dose iterative reconstruction (IR) and \protect\subref{subfig:vivo20ircac} corresponding CAC scoring mask, and \protect\subref{subfig:vivo100fbp} routine-dose FBP reconstruction and \protect\subref{subfig:vivo100fbpcac} corresponding CAC scoring mask. All images have window level/width 90/750 HU. CAC scoring masks show all voxels $\geq 130$ HU in black, and voxels selected by CAC scoring with connected component labeling in red.}
\label{fig:vivoexamples}
\end{figure}

\subsection{Image quality enhancement}
\label{sec:biomedquality}
Medical image acquisition often includes a trade-off between image quality and factors such as time, costs and patient discomfort. For example, lower ionizing radiation dose levels in CT could prevent radiation-induced cancer, but will typically lead to increased image noise levels, and undersampled MR image reconstruction could reduce scan time but may lead to image artifacts. Adversarial methods have been used to avert such effects in the acquisition domain or the image domain. 

To allow CT scanning at low radiation dose, regression CNNs have been proposed to convert low-dose CT images to routine-dose CT images \cite{Kang17}. One problem is that even routine-dose CTs contain low amounts of image noise, and training a regression model can lead to blurred predictions. Wolterink et al. \cite{Wolt17} proposed a 3D model to translate low-dose CT images into routine-dose CT images. The method was evaluated on phantom CT data as well as in-vivo CT images. In phantom CT images, where the low-dose CT and routine-dose CT image were perfectly aligned, the adversarial loss was combined with an $L1$-loss term between the generated image and a reference routine-dose CT image. In real cardiac CT studies, low-dose and routine-dose images were not aligned and self-regularization was used (Eq. \ref{eq:selfreg}). Fig. \ref{fig:vivoexamples} shows a low-dose CT image and the same image denoised using the proposed method or by commercially available iterative reconstruction (IR) software. Both denoising methods compare well with the reference routine-dose CT. However, IR requires CT projection data to be available, while the adversarial method operates on already reconstructed CT images. 

Yang et al. \cite{Yang18c} proposed an alternative adversarial training approach for artifact reduction in low-dose CT. Instead of using the original GAN objective (Eq. \ref{eq:gan}) as in \cite{Wolt17}, Yang et al. proposed to use a Wasserstein distance objective (Eq. \ref{eq:earthmover}) to train the GAN. In addition,  a perceptual loss term based on a pre-trained CNN was added to the loss term. Similarly, Wang et al. \cite{Wang18} proposed to use adversarial methods to synthesize full-dose PET images from low-dose PET images.

In addition to CT artifact reduction, adversarial methods have been found to be useful to speed up MR image acquisition. In compressed sensing MR imaging, the $k$-space is undersampled. The undersampled acquisition can be reconstructed to an MR image, but this image will likely contain aliasing artifacts. In the method proposed by Quan et al. \cite{Quan18}, a generator network tries to transform the reconstructed images into artifact-free MR images. Related methods have been proposed to speed up MR image acquisition with GANs \cite{Mard17,Kim18}, with variations using high-resolution images with different contrasts as additional input alongside the low-resolution image to provide more information to the generator. 

Because the source domain and the target domain are typically strongly related in image quality enhancement problems, it is often sufficient to only train the generator to predict the difference image between the original image and the artifact-free image. This was used by e.g. Wolterink et al. \cite{Wolt17} and Quan et al. \cite{Quan18}. 

\subsection{Image segmentation}
\label{sec:biomedtasks}
Accurate segmentation of anatomical structures is an important topic in medical image analysis and in recent years CNNs have led to many advances in medical image segmentation. One problem when using CNNs is that they are typically trained using a voxel-wise unstructured loss, such as the cross-entropy loss. This may lead to holes and fragments in automatically obtained segmentations. To overcome this, post-processing schemes have been used, such as morphological operations and conditional random fields \cite{KamnCRF}. Alternatively, structurally correct segmentations can be imposed by an adversarial network that assesses whether a segmentation, or a combination of segmentation and input image, is plausible. This approach was successfully employed in medical image segmentation problems. 

Moeskops et al. \cite{Moes17} used an adversarial approach to segment brain tissues in $T_1$-weighted MR brain images. A voxel-wise categorical cross-entropy loss was combined with adversarial feedback from a discriminator network that assessed combinations of images and segmentations. Experiments showed that adversarial training helped prevent segmentation errors and substantially improved Dice similarity indices between automatically obtained and reference segmentations. Moreover, for tumor segmentation in prostate MR Kohl et al. \cite{Kohl17} completely omitted the voxel-wise loss term from the loss function. Hence, in this case optimization was fully driven by the adversarial loss. This resulted in higher Dice similarity indices than training with only  a voxel-wise loss, or training with a combination of voxel-wise loss and adversarial loss. 

One problem of adversarial segmentation methods is that reference segmentations contain discrete label masks, while the generator produces a continuous probability value for each class in each voxel. An adversarial network working directly on reference segmentations and generator outputs could thus learn to distinguish these two by learning to discriminate between discrete and continuous values. One way to overcome this is to let the discriminator look at the product of the input image and the segmentation \cite{Luc16}. Xue et al. \cite{Xue18} proposed to let an adversarial encoder network look at the product of the input image and the predicted segmentation, as well as at the product of the input image and the reference segmentation. Features were extracted at multiple scales for both inputs and the $L1$-loss between the two sets of features served as a scalar adversarial loss for the segmentation CNN, showing to be beneficial for brain tumor segmentation in MR.

Adversarial training could also be used for weakly supervised segmentation of anomalies, in which a label is known for the images but not for individual voxels. Baumgartner et al \cite{Baum17} employed adversarial training with a Wasserstein objective to transform MR images of patients with Alzheimer's disease (AD) into images 
that show what the patient's brain might look like without AD. In this case, the real data distribution contained patients without known AD. The generator was trained to generate a visual attribution map which, when added to the input image, showed a brain without AD. To make sure that the synthesized image matches the anatomical structure in the original image, self-regularization was used (Sec. \ref{sec:cyclegan}). Results showed that obtained attribution maps are more specific than commonly used methods such as class activation mapping.

\subsection{Domain adaptation}
\label{sec:app_dom_adapt}


\begin{figure}[tp]
\centering
\includegraphics[width=0.8\textwidth]{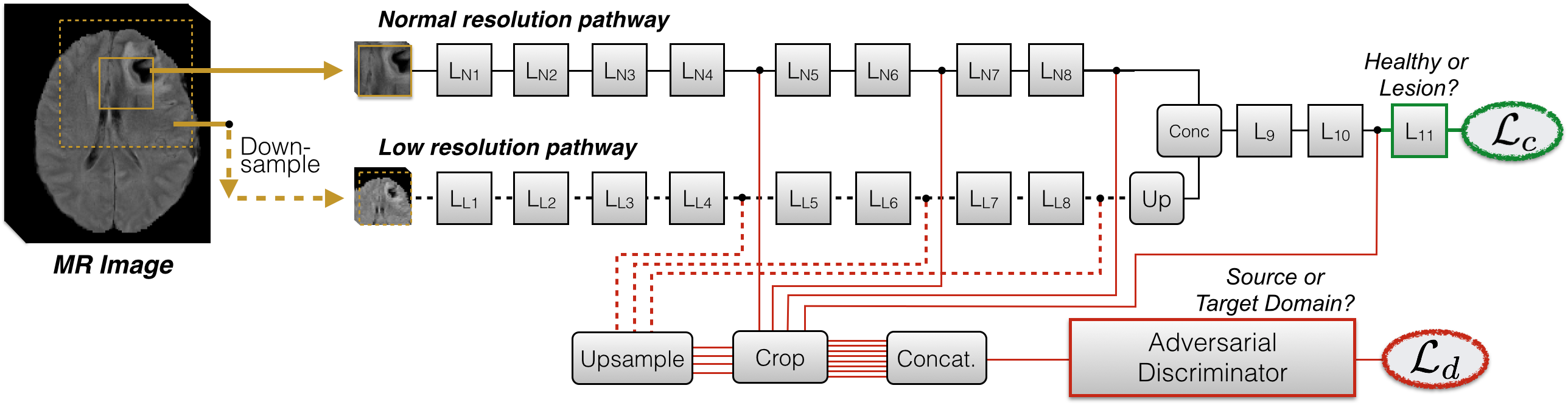}
\caption{Multi-connected adversarial nets for domain adaptation. The domain discriminator processes activations from several depths and scales, which leads to better domain classifier and improved flow of adversarial gradients for better adaptation. Figure adapted from \cite{kamnitsas2017unsupervised}.}
\label{fig:multiconn}
\end{figure}

\begin{figure}[tp]
\centering
\includegraphics[width=0.8\textwidth]{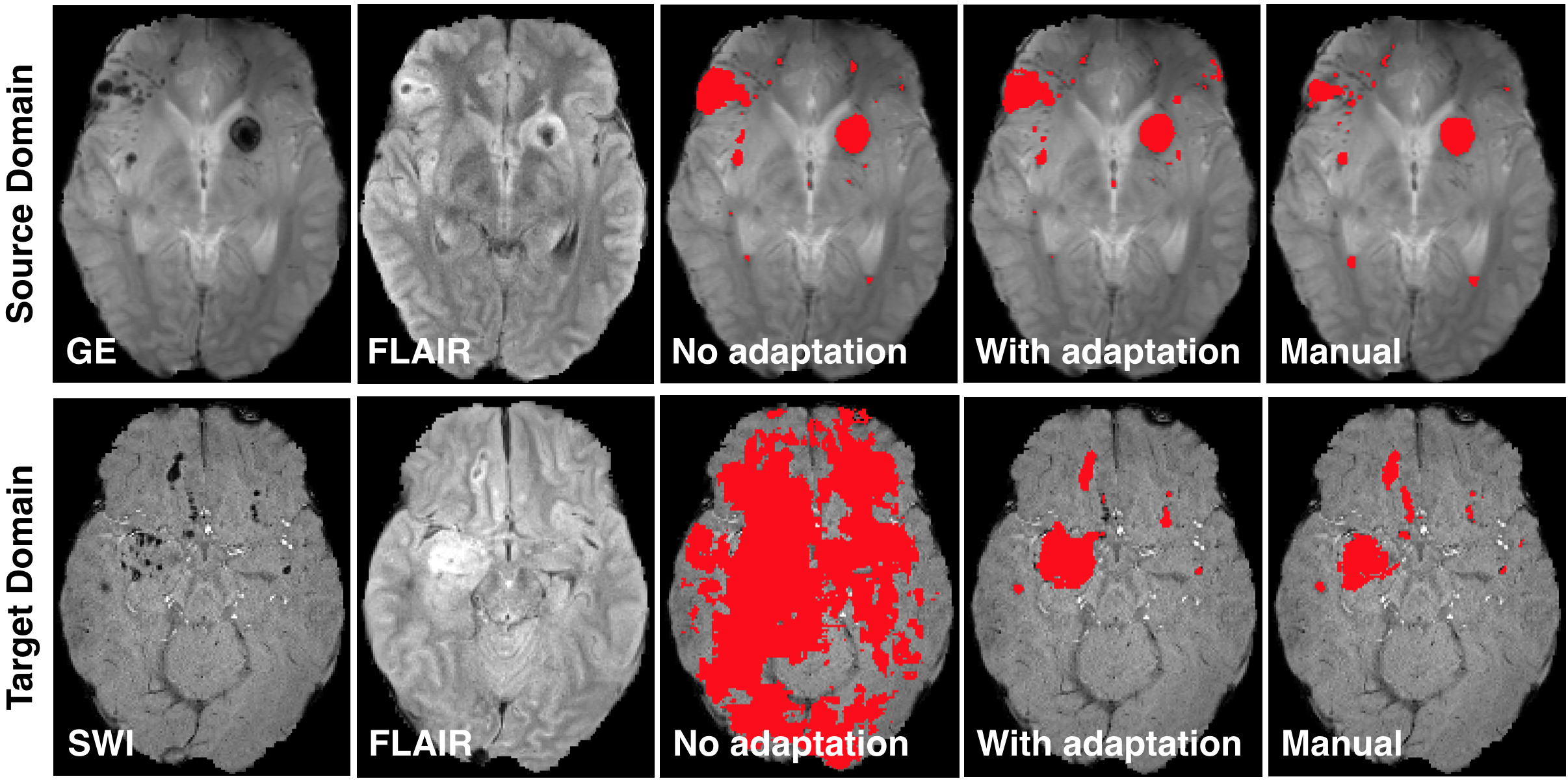}
\caption{Result from Kamnitsas et al \cite{kamnitsas2017unsupervised}. A CNN for segmentation of brain lesions is trained on a database of multi-modal MR scans, which include gradient echo (GE) sequence. The CNN fails when it is applied on another study, where susceptibility weighted imaging was acquired instead of GE. Domain adaptation alleviates the issue.}
\label{fig:uda_visuals}
\end{figure}

The theory and framework for domain adaptation via adversarial training presented in Sec.~\ref{sec:theory_da} has formed the basis for several works in biomedical image analysis. Kamnitsas et al. \cite{kamnitsas2017unsupervised} proposed employing domain adversarial networks for alleviating problematic segmentation due to domain shift between MR acquisition protocols. Extending the basic framework, they proposed multi-connected adversarial nets, which enable the domain discriminator to process information from several layers of the feature extractor (Fig.~\ref{fig:multiconn}). Empirical analysis showed that this leads to a higher quality domain classifier, hence flow of better gradients to the primary network and improved adaptation.
By applying the technique to adapt between two databases of multi-modal MR brain scans with traumatic brain lesions, where one of the modalities differed (Fig.~\ref{fig:uda_visuals}), they showed that domain adversarial training is applicable to 3D CNNs for volumetric image processing. This was previously questioned in the literature \cite{bermudez2016scalable} due to memory constraints.

Recently adversarial networks were employed for adaptation of more segmentation systems. Adaptation with this technique in the case of larger domain shift was recently attempted by Dou et al. \cite{dou2018unsupervised}.
Starting with a segmentation network of the cardiac structures, trained only with labelled MR data, promising results were shown by adapting it to segment CT data, without any labels in CT. In \cite{degel2018domain} the authors investigated the potential of this method for alleviating domain shift in 3D ultrasound between devices of different manufacturers and settings. Adversarial UDA offered significant improvements for the segmentation of the left atrium, which were also complementary to benefits obtained from shape priors \cite{oktay2018anatomically}.

Besides segmentation tasks, Lafarge et al. \cite{lafarge2017domain} investigated the approach in the context of mitosis detection in breast cancer histopathology images. 
As different pathology labs may have slightly different methods for image staining, a model trained on data from one lab may under-perform on data from another. The authors showed that adversarial domain adaptation can offer significant benefits and complement well the more traditional method of color augmentation.

The above works \cite{lafarge2017domain,dou2018unsupervised,degel2018domain} adopted domain discriminators that process information from different depths and scales of the main network. 
This type of domain discriminators was originally proposed by \cite{kamnitsas2017unsupervised} and is appropriate for approaches that learn a domain-invariant \emph{latent space}.


In contrast to the aforementioned approaches, recent works learn mappings between the two domains in \emph{image space}. In Bousmalis et al. \cite{bousmalis2017unsupervised}, simulated images (domain $S$) for which labels are available were mapped to the target domain of real natural images via conditional GANs. A classifier was then trained on the synthetic, labelled, real-looking images, and was afterwards applied on real images. The reverse approach has been shown promising on medical data \cite{mahmood2017unsupervised}. 
A self-regularized conditional GAN mapped real endoscopy images to the domain of simulated images, which were then processed by a predictor trained on simulated data.
The mapping of domains in image space has also been attempted via CycleGANs \cite{huo2018adversarial,chen2018semantic}. In these works, mapping between the domains was also regularized by encouraging semantics to be preserved, so that a segmentor trained with source labels can segment the synthesized images. The approach was found promising in mapping between abdominal MR and CT \cite{huo2018adversarial} and between x-ray scans from different clinical centers \cite{chen2018semantic}.

Learning to map samples between domains in image space offers interpretability in comparison to mapping them in latent space. However, the former assumes no information is exclusive to one domain, otherwise it is an ill-posed problem. In many applications this may not hold. Adaptation in a latent space that only encodes information specific for the primary task (e.g. segmentation) avoids this issue. A more detailed discussion follows in Sec.~\ref{sec:discuss}.

\subsection{Semi-supervised learning}
\label{sec:semisuper}

\begin{figure}[tp]
\centering
\includegraphics[width=0.8\textwidth]{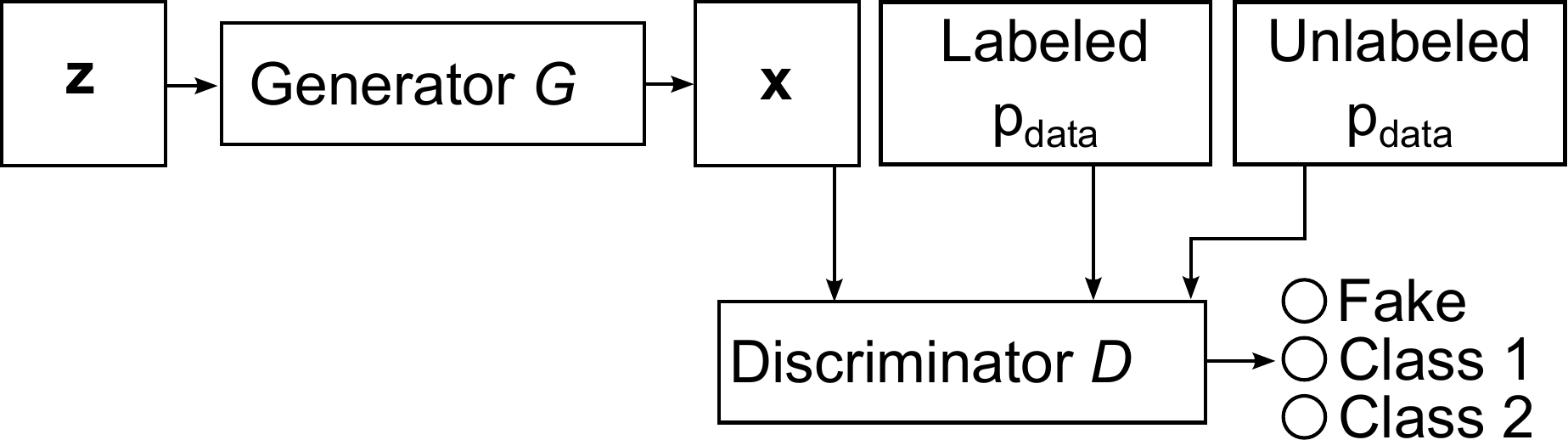}
\caption{Semi-supervised learning with a GAN, as proposed in \cite{Sali16}. The discriminator $D$ receives samples synthesized by $G$, labeled samples from $p_{data}$ that belong to any of a finite number of classes, and samples from the same distribution that belong to a class, but for which the label is not known.}
\label{fig:SGAN}
\end{figure}

Supervised learning methods assume that labels are available for all training samples. In semi-supervised learning (SSL), besides the labeled data, it is assumed that there are also unlabeled data available at training time. The goal of SSL methods is to extract information from the unlabeled data that could facilitate learning a discriminative model with higher performance.

Salimans et al. \cite{Sali16} described a method to use GANs for SSL (Fig. \ref{fig:SGAN}). The discriminator network receives three kinds of samples: samples that have been synthesized by the generator $G$, labeled samples from the data distribution $p_{data}$ that belong to one of a number of classes, and unlabeled samples from the same data distribution. The cost function combines a standard cross-entropy term for the labeled samples with a binary cross-entropy term for the unlabeled and synthetic samples. In other words, the discriminator uses the synthetic and unlabeled samples to learn better feature representations. After training, the discriminator is kept, while the generator is discarded. This method has been applied in
Madani et al. \cite{madani2018semi} for the classification of normal chest x-ray scans versus scans that show cardiac disease. Although the study was performed on limited test data with comparably small networks on downsampled 32x32 scans, it presented preliminary indications that the technique can offer beneficial regularization.

Another approach for SSL with GANs was investigated in Zhang et al. \cite{zhang2017deep}. The method was inspired by the supervised segmentation method of Luc et al. \cite{Luc16} and is also related to Isola et al. \cite{Isol17}. In the latter works, a segmentor is interpreted as the generator of a conditional GAN, producing segmentation maps given an input image, while the discriminator tries to differentiate between predicted and manual segmentations. In Zhang et al. \cite{zhang2017deep} this framework was adapted for SSL. The segmentor is applied on both labeled and unlabeled data. The discriminator of the GAN then tries to distinguish whether a prediction is made on data from the labeled or unlabeled database. 
Adversarial training regularizes the segmentor so that predictions on unlabeled data have similar quality as those on labeled data. This regularization was found beneficial when applied to segmentation of glands in histology images and fungus in electron microscopy \cite{zhang2017deep}.

\section{Discussion and conclusion}
\label{sec:discuss}
Adversarial training is a powerful technique that has helped to further advance deep learning in biomedical image analysis for a large number of tasks. When used for data synthesis, GANs allow the generation of perceptually convincing samples of images or anatomical geometries. The majority of applications of adversarial methods in biomedical image analysis have focused on image synthesis, quality enhancement or segmentation. This has many interesting applications, as shown in the previous section. Despite these benefits, applications of adversarial methods in biomedical image analysis should be critically considered. 

Like all automatic image analysis in medical applications, errors caused by these kinds of methods could have grave clinical consequences. For one, a potential hazard is that generators start ``hallucinating`` content in order to convince the discriminator that their data belongs to the target distribution. This was illustrated in an experiment by Cohen et al. \cite{Cohe18}, who used a set of FLAIR MR brain images \textit{with} tumors and a set of T1-weighted MR brain images \textit{without} tumors to train a CycleGAN. Application of the trained model to a new and unseen FLAIR MR brain image with a tumor led to a perceptually convincing T1-weighted image that nevertheless did not show the tumor. Conversely, a CycleGAN could be trained to synthesize tumors in images of patients that actually did not have tumors. Such side-effects could have dangerous implications, and hence applications should be carefully selected. 
A second problem with CycleGANs arises in the fact that the intermediate representation obtained after application of the first generator network may contain high-frequency information that the second generator network uses to translate the image back to the original domain \cite{Chu17}. It is important to consider that differences exist between real images and synthesized images in the target domain which may not be directly visible. This hidden high-frequency information may affect follow-up analysis with automated methods. For example, a segmentation method that has been trained on real images in the target domain, may have problems dealing with information that the CycleGAN encodes in synthesized images in the target domain.

An important issue for consideration arises when comparing the two approaches presented in Sec.~\ref{sec:app_dom_adapt} for adapting a network to perform a task in two domains. Mapping samples between domains in \emph{image space} instead of \emph{latent space} offers the advantage of interpretability. For example, visual inspection of the synthesized images may reveal failure of the domain transfer and inform that predictions are not trustworthy. 
On the other hand, learning to map samples between domains in image-space requires translating all information present in the images, a broader problem than preserving and translating only the information relevant for the primary task. 
For instance, reconstruction by CycleGANs demands all patterns in the images to be preserved during the transfer from one domain to the other and back. 
This not only requires more complex models \cite{chen2018semantic,huo2018adversarial}, but it may also be an ill-posed problem in some applications. 
This is the case when some information is exclusively present in images from one domain. For instance, certain structures may only be visible in MR but not in CT and vice versa. Enforcing a translation could make the mapping function, such as the generator of a CycleGAN, to ``hallucinate" patterns in order to make the synthetic images look realistic, a behaviour particularly perilous in medical imaging. These issues are mitigated by matching the two distributions in a task-specific \emph{latent space}. This is because the latent space encodes only the information that is relevant for the primary task, which is present in both domains provided that the domains are appropriate for it.

One problem for GANs and adversarial methods is that is challenging to find an appropriate measure for the quality of generated samples. Many works using adversarial techniques aim to generate samples that look ``realistic``, but is is not trivial how this should be measured. Further, it has been shown that low distortion and high perceptual quality are competing goals in image restoration algorithms \cite{Blau18}. 

Future applications of adversarial methods are likely to be found in the same subfields as those described in Sec. \ref{sec:GANSinMIA}. First, de novo generation of new samples with GANs could help enlarge datasets for training of discriminative models. However, this may not directly lead to improved performance of discriminative models trained using this data. It is possible that there is no additional information in the synthesized samples beyond that which is already present in the training data set. Furthermore, despite recent advances in the training of GANs (Sec. \ref{sec:objectives}), successfully training a GAN to generate perceptually convincing samples can be challenging and further advances are needed in this respect. Second, we expect that there will be many applications for conversion of images from one modality to another modality, for removal of artifacts, or for image segmentation. This may lead to a decrease in the number of medical images that are being acquired, which could benefit patients directly. Additionally, acquisition of images with lower radiation dose or sparse sampling of $k$-space could directly affect patients. Finally, future applications of adversarial methods may focus on the identification of abnormalities. Given that a large numbers of images are available for healthy patients, models could be trained to learn the distribution of healthy patients and to identify patients with disease as deviating from this distribution.

We conclude that -- when used with caution -- adversarial techniques have many potential applications in medical imaging. Adding adversarial feedback to a model performing tasks such as segmentation or image synthesis, for which it is hard to hand-craft a loss function, will in many cases yield images that more closely match real images in the target domain and are more likely to be valuable in clinical practice.

\bibliographystyle{elsarticle-num} 
\bibliography{literature}

\end{document}